\documentclass[letterpaper]{article} 
\usepackage{aaai2026}  
\usepackage{times}  
\usepackage{helvet}  
\usepackage{courier}  
\usepackage[hyphens]{url}  
\usepackage{graphicx} 
\usepackage{natbib}  
\usepackage{caption} 
\usepackage{algorithm}
\usepackage{algorithmic}
\usepackage{cite}
\usepackage{amssymb}
\usepackage{booktabs}
\usepackage{multirow}
\usepackage{amsmath} 
\usepackage{makecell}
\newcommand\xrowht[2][0]{\addstackgap[.5\dimexpr#2\relax]{\vphantom{#1}}}
\usepackage{stackengine}
\usepackage{newfloat}
\usepackage{listings}
\usepackage{tabularx}
\DeclareCaptionStyle{ruled}{labelfont=normalfont,labelsep=colon,strut=off} 
\floatstyle{ruled}
\newfloat{listing}{tb}{lst}{}
\floatname{listing}{Listing}
\title{APT: Affine Prototype-Timestamp For Time Series Forecasting \\Under Distribution Shift}
\author {
	Yujie Li,\textsuperscript{\rm 1,2}
	Zezhi Shao,\textsuperscript{\rm 1}
	Chengqing Yu,\textsuperscript{\rm 1,2}
	Yisong Fu,\textsuperscript{\rm 1,2}
	Tao Sun,\textsuperscript{\rm 1}
	Yongjun Xu,\textsuperscript{\rm 1,2}
	Fei Wang\textsuperscript{\rm 1,2}\thanks{Corresponding authors.}
}
\affiliations{
	\textsuperscript{\rm 1}State Key Laboratory of AI Safety, Institute of Computing Technology,Chinese Academy of Sciences\\
	\textsuperscript{\rm 2}University of Chinese Academy of Sciences\\
	\{liyujie23s, shaozezhi, yuchengqing22b, fuyisong24s, suntao, xyj, wangfei\}@ict.ac.cn
}

\usepackage{bibentry}

\begin{document}
	
	\maketitle
	
	\begin{abstract}
		Time series forecasting under distribution shift remains challenging, as existing deep learning models often rely on local statistical normalization~(e.g., mean and variance) that fails to capture global distribution shift. Methods like RevIN and its variants attempt to decouple distribution and pattern but still struggle with missing values, noisy observations, and invalid channel-wise affine transformation. To address these limitations, we propose \textbf{A}ffine \textbf{P}rototype-\textbf{T}imestamp~(\textbf{APT}), a lightweight and flexible plug-in module that injects global distribution features into the normalization–forecasting pipeline. By leveraging timestamp-conditioned prototype learning,
		APT dynamically generates affine parameters that modulate both input and output series, enabling the backbone to learn from self-supervised, distribution-aware clustered instances.
		APT is compatible with arbitrary forecasting backbones and normalization strategies while introducing minimal computational overhead. Extensive experiments across six benchmark datasets and multiple backbone-normalization combinations demonstrate that APT significantly improves forecasting performance under distribution shift.
	\end{abstract}

\begin{links}
	\link{Code}{https://github.com/blisky-li/APT}
\end{links}

\section{Introduction}
Time series reflect the artificial or natural regularities of complex dynamic systems across transportation~\cite{li2024dynamic, shao2022spatial}, health~\cite{ferte2024reservoir} and weather~\cite{yu2025ginar+}. However, external factors commonly induce distributional shift and non-stationarity, making forecasting models struggle to effectively capture patterns under changing statistical properties.

Reversible Instance Normalization~(RevIN)~\cite{kim2021reversible} introduces a two-stage paradigm for mitigating temporal distribution shift: instance normalization that removes instance-specific distributions and affine transformation that attempts to restore channel-wise distributional features.
This decoupling of time-varying distributions from learnable temporal patterns has laid the foundation for many modern forecasting models, and subsequent advancements—including DishTS~\cite{fan2023dish}, SAN~\cite{liu2023adaptive}, SIN~\citep{han2024sin}, and FAN~\cite{DBLP:conf/nips/YeDZG24}—have further refined this framework by modeling future distributions adaptively to address intra-instance shift.

\begin{figure}[t]
	\centering
	\includegraphics[width=0.40\textwidth]{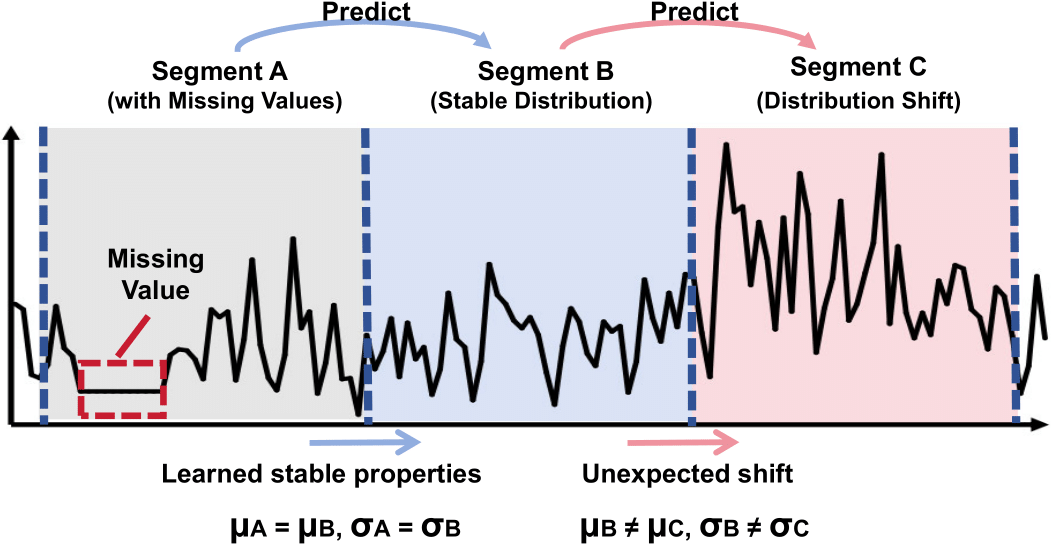}
	\caption{Local statistical normalization fails to handle distribution shift on ECL. The model retains outdated statistics when predicting from flawed Segment A to B but faces unseen shifts in Segment B and C. Shifts across all three segments exceed prior inter and intra shift issues, posing a broader global shift challenge for APT.}
	\label{chal}
\end{figure}

Despite their success, RevIN-like methods still face inherent limitations. First, Local statistics such as mean and variance are sensitive to missing values and noise, which are common in time series. Second, their instance-specific normalization fails to capture global shifts, as shown in Figure~\ref{chal}, where features learned from Segments A and B collapse in Segment C under abrupt change.

Moreover, the channel-wise static affine transformation~\citep{bebis1999learning} in RevIN yields only limited improvements as shown in Table~\ref{affine1}, and is entirely omitted in follow-up works such as DishTS and SAN.
This suggests static transformations fail to address inter-channel distribution variance and are ineffective against distribution shift.

\begin{table*}[t]
	\centering 		 		 		
	
	\small
	\renewcommand\arraystretch{0.95}
	\setlength{\tabcolsep}{1.5pt}{
		\begin{tabular}{c|cc|cc|cc|cc|cc|cc|cc|cc|cc}
			\toprule[1.2pt]
			\multicolumn{1}{c|}{Methods} & \multicolumn{4}{c|}{CATS} & \multicolumn{4}{c|}{Informer+RevIN} & \multicolumn{4}{c|}{iTransformer} & \multicolumn{4}{c|}{SparseTSF} & \multicolumn{2}{c}{Avg. $\Delta$}\\
			\midrule
			{Affine~(RevIN)} & \multicolumn{2}{c|}{True} & \multicolumn{2}{c|}{False}& \multicolumn{2}{c|}{True} & \multicolumn{2}{c|}{False}& \multicolumn{2}{c|}{True} & \multicolumn{2}{c|}{False}& \multicolumn{2}{c|}{True} & \multicolumn{2}{c|}{False}& \multicolumn{2}{c}{True - False}\\
			\midrule
			\multicolumn{1}{c|}{Metric} & MAE & MSE & MAE & MSE& MAE & MSE& MAE & MSE& MAE & MSE& MAE & MSE& MAE & MSE& MAE & MSE& MAE & MSE\\
			\midrule[1pt]
			\xrowht{1pt}
			ECL& 
			\textbf{0.260} & \textbf{0.165} & 0.261 & \textbf{0.165} 
			& \textbf{0.307} & \textbf{0.209} & 0.308 & 0.212 
			& \textbf{0.264} & \textbf{0.169} & \textbf{0.264} & 0.170 
			& 0.266 & 0.173 & \textbf{0.265} & \textbf{0.171} & 0.000 & $\downarrow$ 0.001\\
			\xrowht{8pt}
			Weather & 
			\textbf{0.274} & \textbf{0.236} & \textbf{0.274} & 0.241 
			& 0.311 & 0.307 & \textbf{0.310} & \textbf{0.306} 
			& 0.284 & 0.251 & \textbf{0.282} & \textbf{0.248} 
			& \textbf{0.294} & \textbf{0.267} & 0.297 & 0.269 & 0.000 & $\downarrow$ 0.001\\
			\xrowht{8pt}
			ETTh1& 
			0.447&0.456&\textbf{0.441}&\textbf{0.451}
			&\textbf{0.582} & 0.681 & 0.584 & \textbf{0.679} 
			& \textbf{0.462} & \textbf{0.471} & 0.464 & 0.475 
			& 0.440 & 0.445 & \textbf{0.430} & \textbf{0.437} &$\uparrow$ 0.003& $\uparrow$ 0.003\\
			\xrowht{8pt}
			Exchange& 
			0.437 & 0.352 & \textbf{0.436} & \textbf{0.349} 
			& \textbf{0.677} & 0.708 & \textbf{0.644} & 0.727
			& 0.493 & 0.423 & \textbf{0.475} & \textbf{0.398} 
			& 0.499 & 0.434 & \textbf{0.492} & \textbf{0.428} & $\uparrow$0.015 & $\uparrow$0.004\\
			\bottomrule[1.2pt]
	\end{tabular}}
	\caption{The affine transformations in RevIN do not provide performance gains, Metric $\uparrow$ means worse performance.}
	\label{affine1}
\end{table*}

To overcome the limitations of local statistical normalization, we propose \textbf{A}ffine \textbf{P}rototype-\textbf{T}imestamps~(APT), a lightweight and model-agnostic plug-in. APT replaces static affine transformations in standard normalization with dynamically generated parameters conditioned on timestamps, inspired by vision style transfer~\citep{huang2017arbitrary}.
This enables the forecasting pipeline to access global temporal semantics, which serve as an underlying schedule guiding the behavior of the time series system, thereby enhancing robustness and adaptability under distribution shift.

\begin{figure}[h]
	\centering
	\includegraphics[width=0.40\textwidth]{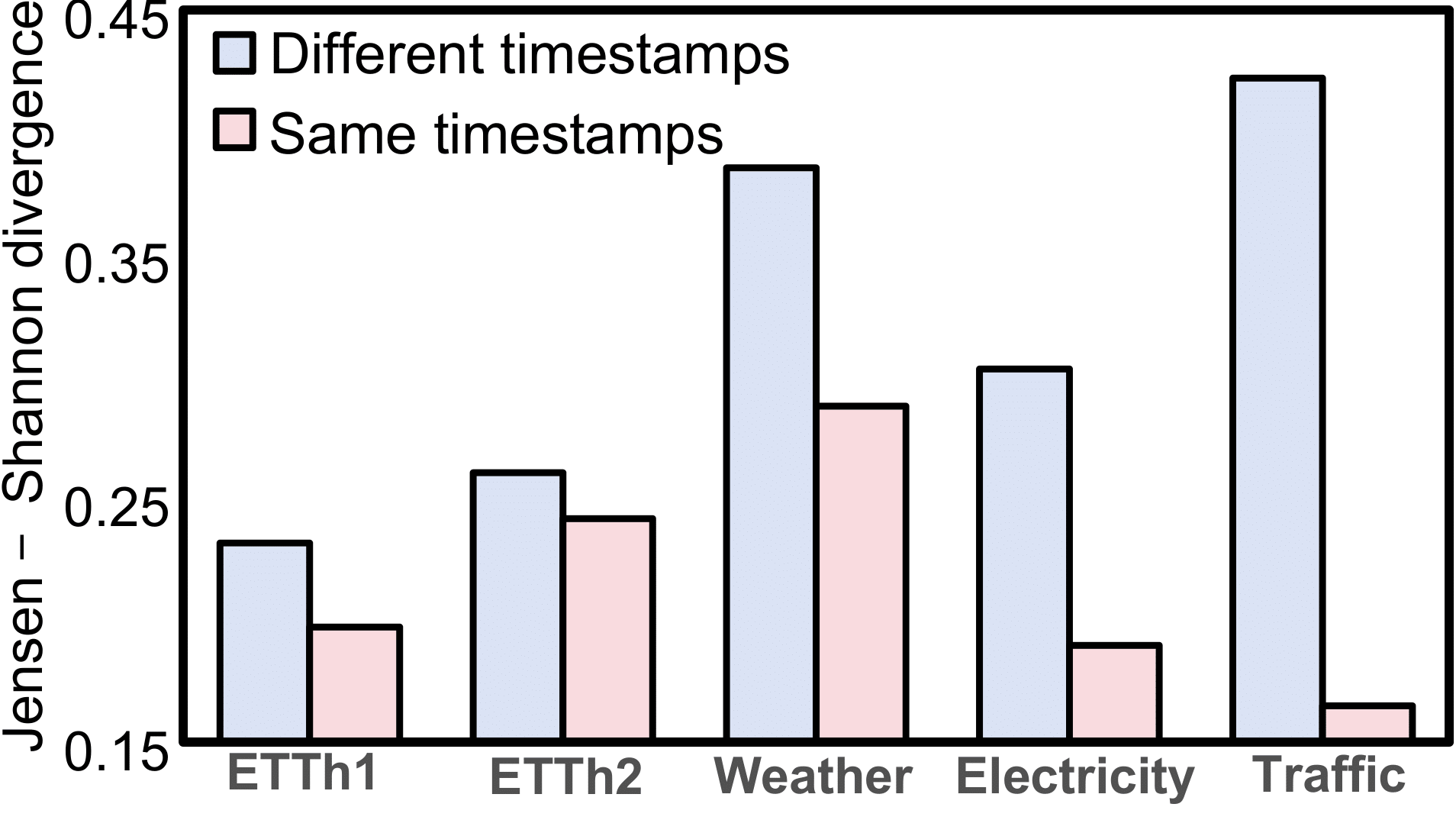}
	\caption{JS divergence of subseries with the same timestamp label in benchmark datasets}
	\label{js}
\end{figure}

Easily accessible timestamps are widely regarded as containing global temporal semantics~\cite{li2025sta},
and our empirical analysis in Figure~\ref{js} confirms subseries sharing similar timestamp labels such as "Time in Day" and "Day in Week" tend to exhibit distributional similarity. APT exploits this regularity by embedding timestamps into a shared latent space to enable conditioned affine transformations.

However, the sparsity of fine-grained timestamps combinations can cause insufficient training or poor generalization to unseen cases. 
Inspired by few-shot learning~\cite{li2022prototype}, APT employs prototype learning, replacing raw timestamp embeddings with nearest-neighbor prototypes to yield more robust representations. To further promote diversity and balanced prototype usage, we introduce orthogonality and load-balancing losses.

Rather than learning static affine parameters, APT utilizes global temporal semantics in timestamps to adaptively assign distinct affine parameters to different subseries. To achieve this, it first encodes timestamps and obtains robust representations through prototype matching, then converts prototype embeddings into low-dimensional affine parameters via MLPs. Crucially, this process employs self-supervised learning with the backbone and normalization frozen, leveraging orthogonal loss for embedding diversity, load-balancing loss for prototype usage, and extra affine regularization loss to ensure affine parameter convergence. 

In summary, the central function of APT is to supply forecasting backbones with learnable and global distribution features suppressed by normalization or disrupted by distribution shift; all components and optimization of APT are explicitly designed to serve this purpose.

Our contributions are as follow:

\begin{itemize}
	\item We propose APT, a lightweight and model-agnostic plug-in for overcoming limitations of local statistical normalization and mitigating distribution shift.
	\item APT generates dynamic affine parameters via timestamp-conditioned prototype matching and self-supervised learning, delivering learnable distribution features compatible with any forecasting backbone and normalization.
	\item Extensive experiments with diverse backbones and normalization methods confirm APT's effectiveness in improving forecasting accuracy under distribution shift. 
\end{itemize}

\section{Related Work}
\subsection{Deep Time Series Forecasting}
As the backbone of forecasting, time series models are primarily designed to learn temporal patterns such as seasonality and trends~\cite{wang2025aries, shao2025blast}. Deep models like Informer~\cite{zhou2021informer} improve long-term forecasting, while DLinear~\cite{zeng2023transformers} shows that simple MLP can also perform well. Recent developments in forecasting focus on complex temporal dependencies, including intra-channel long-term dependencies, as explored by PatchTST~\cite{nietime} and SparseTSF~\cite{linsparsetsf}, as well as inter-channel interactions, as addressed by iTransformer~\cite{liuitransformer} and CATS~\cite{DBLP:conf/nips/Kim00K24}.

Distribution shift is not typically the focus of forecasting backbone, but it is particularly important in the forecasting pipeline because it significantly increases the difficulty of pattern learning and reduces forecasting performance.

\subsection{Normalization for Distribution Shift}
Normalization has become a standard component in time series forecasting due to its efficiency in handling distribution shift.
RevIN~\cite{kim2021reversible} proposes a distribution-pattern decoupling scheme by forcing the normalization of histories to a unified domain and de-normalizing predictions, which greatly reduces the complexity of non-stationary forecasting~\cite{fu2025selective}.

Subsequent works aim to recover suppressed distributional features because the distribution of history and future within a single instance is inconsistent. DishTS~\cite{fan2023dish} learns intra- and inter-segment drift. SAN~\cite{liu2023adaptive} focuses on patch-level shift. SIN~\cite{han2024sin} relearns statistics based on local invariance and global variability criteria, and FAN~\cite{DBLP:conf/nips/YeDZG24} replaces normalization with main frequency extraction and residual forecasting.

\begin{figure*}[t]
	\centering
	\includegraphics[width=0.88\textwidth]{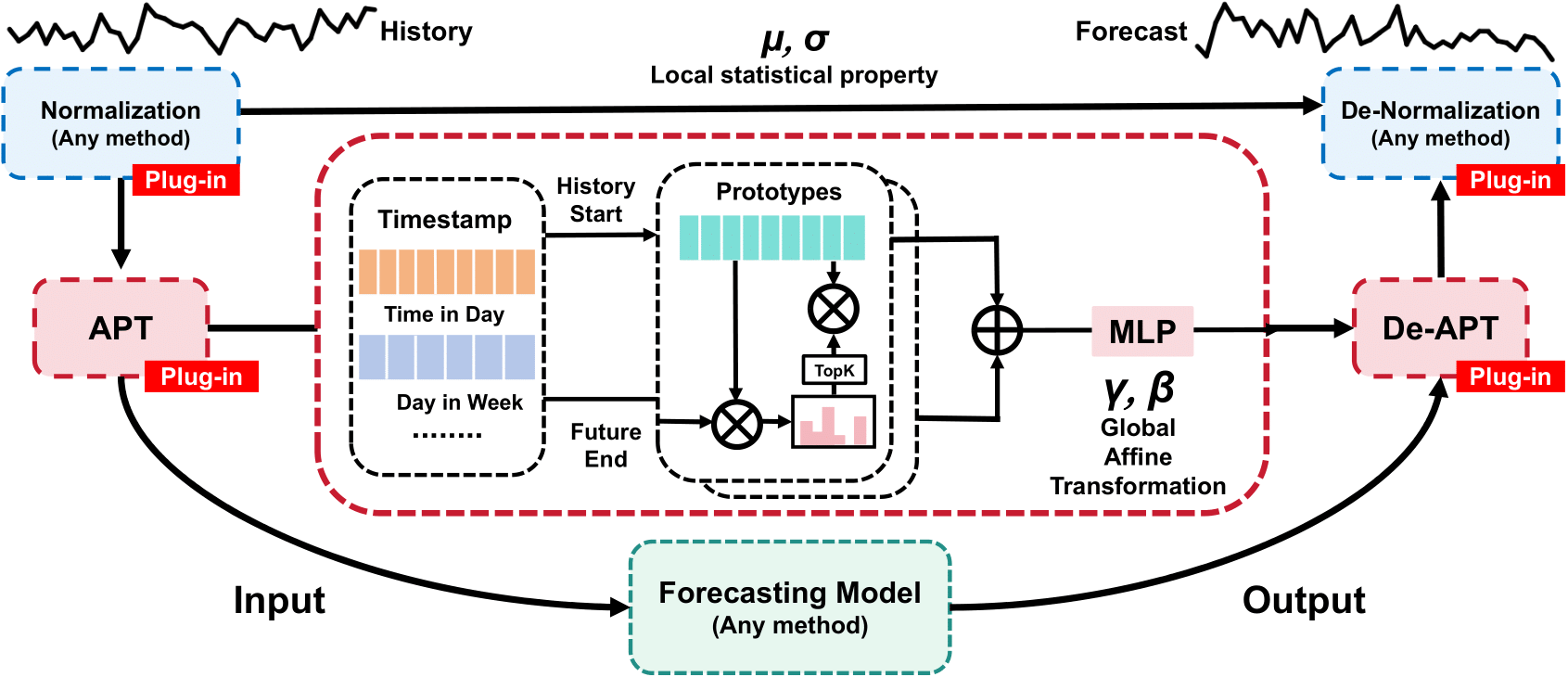}
	\caption{The pipeline of time series forecasting and the schematic of APT.}
	\label{aptpipeline}
\end{figure*}

\subsection{Affine Transformation}
Affine transformations were initially used to align parameter spaces between layers after normalization, improving training stability of deep models. In conditional generative networks, they serve to reintroduce task-specific information by modulating distributional features~\cite{karras2019style}.  
CIN~\cite{dumoulin2017learned} assigns affine parameters to each style via instance normalization. AdaIN~\cite{huang2017arbitrary} extracts them from target images for arbitrary style transfer, and FiLM~\cite{perez2018film} extends this idea to feature-level conditioning. Such conditional affine modulation has become standard across vision and language tasks~\cite{ziegler2019encoder}.

However, such conditional normalization techniques are rarely explored in time series forecasting, where normalization still mainly depends on local statistics. Inspired by these advances, APT introduces timestamp-based conditioning to generate dynamic affine parameters, enabling better adaptation to distribution shift.

\section{Preliminaries}
\subsection{Time Series Forecasting}
Given the historical multivariate time series $\{x_{t-L:t}^{(i)}\}_{i=1}^{C} = \{x_{t-L:t}^{(1)}, ..., x_{t-L:t}^{(C)}\}\in \mathbb{R}^{L \times C}$,
where $x_{t}$ refer to the values at timestamp $t$, $L$ is the length of historical window and $C$ is the number of channels. The objective of time series forecasting is to predict future series $\{y_{t+1:t+H}^{(i)}\}_{i=1}^{C} = \{y_{t+1:y+H}^{(1)}, ..., y_{t+1:y+H}^{(C)}\}\in \mathbb{R}^{H \times C}$ by leveraging the forecasting model $\boldsymbol{\mathcal{M}}$ to learn patterns from historical data, where $H$ is the future horizon:
\begin{equation}
	\{y_{t+1:t+H}^{(i)}\}_{i=1}^{C} = \boldsymbol{\mathcal{M}}(\{x_{t-L:t}^{(i)}\}_{i=1}^{C})
\end{equation}

\subsection{Time Series Normalization}
Given a normalization method $\boldsymbol{\mathcal{N}}$, it normalizes historical time series before inputting them into $\boldsymbol{\mathcal{M}}$ to eliminate distribution differences, and then de-normalizes the time series output by $\boldsymbol{\mathcal{M}}$ to restore distribution information:
\begin{equation}\label{normal}
	\{y_{t+1:t+H}^{(i)}\}_{i=1}^{C} = \boldsymbol{\mathcal{N}}^{-1}(\boldsymbol{\mathcal{M}}(\boldsymbol{\mathcal{N}}(\{x_{t-L:t}^{(i)}\}_{i=1}^{C})))
\end{equation}

Since statistics often have additional networks in $\boldsymbol{\mathcal{N}}$ for adaptive learning, the statistics adopted by $\boldsymbol{\mathcal{N}}$ and $\boldsymbol{\mathcal{N}}^{-1}$ often differ from the originals and are not identical to each other. Take mean-variance normalization as an example. $\mu_{t}$ is the mean of historical time series $\{x_{t-L:t}^{(i)}\}_{i=1}^{C}$, and $\sigma_{t}$ is its variance.  The subscripts $l$ and $h$ donate the normalization and de-normalization stages respectively, where the statistics are either preserved~(RevIN) or relearned~(Dish-TS, SAN), depending on the normalization strategy. Equation~\ref{normal} can be expressed as:
\begin{equation}\label{normal2}
	y_{t+1:t+H}^{(i)} = \sigma_{h, t}^{(i)}\boldsymbol{\mathcal{M}}(\frac{x_{t-L:t}^{(i)} - \mu_{l, t}^{(i)}}{\sigma_{l, t}^{(i)}}) + \mu_{h, t}^{(i)}
\end{equation}

\section{Methodology}
\subsection{Overview}
In Figure~\ref{aptpipeline}, the mainstream paradigm in time series forecasting follows a normalization–forecasting–denormalization pipeline. In this framework, normalization modules~(e.g., RevIN, SAN) serve as plug-in components to either remove or learn distributional information, while forecasting models~(e.g., SparseTSF, iTransformer) constitute the backbone for capturing temporal patterns.

Affine Prototype-Timestamp~(APT) is a lightweight plug-in module designed to break through the limitations of local statistical normalization strategies. Within the forecasting pipeline, the forward transformation of APT is applied after normalization and before model input, and the inverse transformation~(de-APT) is applied after model output and before de-normalization.

Instead of learning a single set of affine parameters $\gamma$ and $\beta$, APT leverages global timestamps to adaptively assign distinct affine parameters to different subseries. The goal is to reintroduce learnable distributional features, conditioned on global constraints, after distributional features have been suppressed by normalization or disrupted due to distribution shift. The following sections detail the implementation and training procedure of APT.

\subsection{Affine Prototype-Timestamp}
To address global distribution shift and reduce reliance on local statistical normalization, we propose APT, a module that learns adaptive affine parameters conditioned on timestamp representations. Timestamps, commonly formatted as \%Y-\%m-\%d \%H:\%M, are widely available and encode rich temporal semantics, making them a natural proxy for global distributional regularities.

Given the input historical sequence $\{x_{t-L:t}^{(i)}\}_{i=1}^{C}$ and forecast horizon $H$, we extract the start timestamp $ts_{t-L}$ of the history and the end timestamp $ts_{t+H}$ of the forecasting horizon for each sample. To obtain semantically meaningful representations, we discretize timestamps into categorical attributes, such as “Time in Day” ($\mathbf{TiD}$) and “Day in Week” ($\mathbf{DiW}$), which are broadly applicable and empirically correlated with global temporal variation.

Each timestamp $t$ is represented by the sum of its corresponding attribute embeddings $\textbf{\textsl{T}}_{t}$:
\begin{equation}\label{id}
	\textbf{\textsl{T}}_{t} = \textbf{\textsl{T}}^{\mathrm{TiD}}_{t} + \textbf{\textsl{T}}^{\mathrm{DiW}}_{t}
\end{equation}

However, discrete timestamp combinations can be sparse and fail to generalize across unseen scenarios. Inspired by few-shot learning, we replace raw timestamp embeddings with learnable prototypes that capture shared temporal semantics.
As a result, we match each timestamp embedding $\textbf{\textsl{T}}_{t} \in \mathbb{R}^{D}$ against a shared learnable prototype library $\textbf{\textsl{P}}=\{\textbf{\textsl{p}}_{j} \in \mathbb{R}^{D}\}_{j=1}^{N}$ via inner product similarity:
\begin{equation}
	\textbf{\textsl{S}}_{t} = \textbf{\textsl{T}}_{t} \textbf{\textsl{P}}^\top
\end{equation}

We then select the top‑$\mathit{k}$ most similar prototypes per timestamps and compute their softmax similarity scores:
\begin{equation}\label{softmax}
	\textbf{\textsl{w}}_{t, [1:k]} = \mathrm{Softmax}(\textbf{\textsl{S}}_{t, [1:k]})
\end{equation}

This yields a sparse weight matrix $\textbf{\textsl{W}}_{t} \in \mathbb{R}^{N}$, where only the top‑$\mathit{k}$ entries per row are non-zero. The final timestamp representation is formed via a weighted aggregation over the prototype embeddings:
\begin{equation}\label{w}
	\widetilde{\textbf{\textsl{T}}}_{t} = \sum_{j=1}^{N} \textbf{\textsl{W}}_{t, j} \textbf{\textsl{p}}_{j}
\end{equation}
To encode both past and future temporal context, we aggregate their embeddings: $\widetilde{\textbf{\textsl{T}}}_{t} = \widetilde{\textbf{\textsl{T}}}_{t}^{history} + \widetilde{\textbf{\textsl{T}}}_{t}^{futher}$ . Optionally, channel identity embeddings $\textbf{\textsl{ID}}\in \mathbb{R}^{C \times D}$ can be added to incorporate per-channel variation when necessary.

Next, we employ two multi-layer perceptrons to map the aggregated embedding to channel-wise affine parameters $\gamma, \beta \in \mathbb{R}^{\{C\} \times 1}$, where $\{C\}$ denotes the option of whether to adopt identity embeddings:
\begin{equation}\label{idemb}
	\gamma_{t}, \beta_{t} = \mathrm{MLP}_{\gamma}(\widetilde{\textbf{\textsl{T}}}_{t}),  \mathrm{MLP}_{\beta}(\widetilde{\textbf{\textsl{T}}}_{t})
\end{equation}

These learned parameters modulate the normalized time series before and after the forecasting model. Substituting them into the normalization pipeline in Equation~\ref{normal2}, the revised inference process becomes:
\begin{equation}\label{normal3}
	y_{t\!+\!1:t\!+\!H}^{(i)} \!=\! \frac{\sigma_{h, t}^{(i)}}{\gamma_{t}^{(i)}}\! (\!\boldsymbol{\mathcal{M}}(\gamma_{t}^{(i)}\frac{x_{t-L:t}^{(i)} \!-\! \mu_{l, t}^{(i)}}{\sigma_{l, t}^{(i)}} \!+\! \beta_{t}^{(i)}\!) \!-\! \beta_{t}^{(i)}\!) \!+\! \mu_{h, t}^{(i)}
\end{equation}

This formulation enables globally informed affine transformations that better adapt to distributional variations across time.

\begin{table*}[t]
	\centering 		 		 		
	\small
	\renewcommand\arraystretch{0.88}
	\label{maintable}
	\setlength{\tabcolsep}{3pt}{
		\begin{tabular}{c|c|cc|cc|cc|cc|cc|cc|cc|cc}
			\toprule[1.2pt]
			\multicolumn{2}{c|}{\textbf{Methods}}  & \multicolumn{4}{c|}{\textbf{CATS}} & \multicolumn{4}{c|}{\textbf{Informer}} & \multicolumn{4}{c|}{\textbf{iTransformer}} & \multicolumn{4}{c}{\textbf{SparseTSF}}\\ 
			\midrule
			\multicolumn{2}{c|}{\textbf{Affine}} & && \multicolumn{2}{c|}{\textbf{+APT}}&&& \multicolumn{2}{c|}{\textbf{+APT}}&&& \multicolumn{2}{c|}{\textbf{+APT}}&&& \multicolumn{2}{c}{\textbf{+APT}}\\
			\midrule
			\multicolumn{2}{c|}{\textbf{Metrics}} & MAE & MSE & MAE & MSE& MAE & MSE& MAE & MSE& MAE & MSE& MAE & MSE& MAE & MSE& MAE & MSE\\
			\midrule[1.2pt]
			\multirow{5}{*}{\rotatebox[origin=c]{90}{\textbf{ECL}}} & \textbf{None} &0.264&0.168&\textbf{0.262}&\textbf{0.166}&0.426&0.369&\textbf{0.408}&\textbf{0.332}
			&0.270&0.169&\textbf{0.265}&\textbf{0.165}&0.269&0.173&\textbf{0.267}&\textbf{0.171}
			\\
			&\textbf{+RevIN}&\textbf{0.258}&\textbf{0.165}&0.259&0.166&0.337&0.252&\textbf{0.326}&\textbf{0.233}
			&\textbf{0.259}&0.164&\textbf{0.259}&\textbf{0.162}&0.265&\textbf{0.173}&\textbf{0.264}&\textbf{0.173}
			\\
			&\textbf{+Dish-TS}&0.272&0.173&\textbf{0.270}&\textbf{0.172}&0.379&0.305&\textbf{0.371}&\textbf{0.297}&0.266&0.166&\textbf{0.261}&\textbf{0.161}&0.272&0.173&\textbf{0.267}&\textbf{0.171}
			\\
			&\textbf{+SAN}&0.277&\textbf{0.175}&\textbf{0.276}&\textbf{0.175}&0.349&0.266&\textbf{0.324}&\textbf{0.231}
			&\textbf{0.258}	&	\textbf{0.159}	&	0.261	&	0.160
			&0.271&0.172&\textbf{0.266}&\textbf{0.165}\\
			&\textbf{+FAN}&0.265&0.167&\textbf{0.263}&\textbf{0.165}
			&0.266&\textbf{0.165}&\textbf{0.264}&0.166&0.274&0.172&\textbf{0.268}&\textbf{0.169}
			&0.263&0.166&\textbf{0.261}&\textbf{0.164}
			\\ 
			\midrule
			\multirow{5}{*}{\rotatebox[origin=c]{90}{\textbf{ETTh1}}} & \textbf{None} &0.469&0.486&\textbf{0.436}	&	\textbf{0.431}&
			0.891&1.255&\textbf{0.783}	&	\textbf{1.134}
			&0.499&0.503&\textbf{0.463}	&	\textbf{0.457}
			&0.439&0.436&\textbf{0.437}	&	\textbf{0.435}\\
			&\textbf{+RevIN}&0.443&0.448&\textbf{0.431}	 &	\textbf{0.427}
			&0.597&0.715&\textbf{0.562}	&	\textbf{0.657}&0.473&0.485&\textbf{0.441}	&	\textbf{0.441}
			&0.429&0.427&\textbf{0.427}	&	\textbf{0.424}
			\\
			&\textbf{+Dish-TS}&0.475&0.485&\textbf{0.452}	&	\textbf{0.456}&0.797&1.062&\textbf{0.721}	&	\textbf{0.947}&0.498&0.513&\textbf{0.492}	&	\textbf{0.495}
			&0.467&0.477&\textbf{0.456}	&	\textbf{0.461}
			\\
			&\textbf{+SAN}&0.451&0.472&\textbf{0.447}	&	\textbf{0.463}
			&0.542&0.614&\textbf{0.528}	&	\textbf{0.599}&0.466&0.477&\textbf{0.453}	&	\textbf{0.470}
			&0.504&0.564&\textbf{0.464}	&	\textbf{0.488}\\
			&\textbf{+FAN}&0.478&0.482&\textbf{0.473}	&	\textbf{0.478}&0.519&0.533&\textbf{0.506}	&	\textbf{0.529}
			&0.487&0.497&\textbf{0.477}	&	\textbf{0.480}
			&0.484&0.497&\textbf{0.480}	&	\textbf{0.488}\\ 
			\midrule
			\multirow{5}{*}{\rotatebox[origin=c]{90}{\textbf{ETTh2}}} & \textbf{None} &0.504&0.533&\textbf{0.471}	&	\textbf{0.486}&1.433&2.883&\textbf{1.194}	&	\textbf{2.269}
			&0.686&0.863&\textbf{0.493}	&	\textbf{0.507}
			&0.492&0.515&\textbf{0.484}	&	\textbf{0.498}
			\\
			&\textbf{+RevIN}&0.425&0.392&\textbf{0.417}	&	\textbf{0.390}
			&0.565&0.658&\textbf{0.504}	&	\textbf{0.531}
			&0.439&0.421&\textbf{0.422}	&	\textbf{0.393}&\textbf{0.427}	&	\textbf{0.399}	&	0.428	&	0.470
			\\
			&\textbf{+Dish-TS}&0.476&0.466&\textbf{0.444}	&	\textbf{0.419}&1.321&3.261&\textbf{0.795}	&	\textbf{1.318}
			&0.537&0.587&\textbf{0.485}	&	\textbf{0.497}
			&0.631&0.933&\textbf{0.533}	&	\textbf{0.612}
			\\
			&\textbf{+SAN}&0.438&0.406&\textbf{0.433}	&	\textbf{0.404}&0.718&0.939&\textbf{0.447}	&	\textbf{0.438}
			&0.438&0.418&\textbf{0.414}	&	\textbf{0.410}
			&0.559&0.619&\textbf{0.485}	&	\textbf{0.478}\\
			&\textbf{+FAN}&0.477&0.487&\textbf{0.470}	&	\textbf{0.470}
			&0.565&0.624&\textbf{0.489}	&	\textbf{0.499}&0.492&0.512&\textbf{0.483}	&	\textbf{0.497}
			&0.475&\textbf{0.467}	&	\textbf{0.472}	&	0.473\\ 
			\midrule
			\multirow{5}{*}{\rotatebox[origin=c]{90}{\textbf{Exchange}}} & \textbf{None} &0.617&0.888&\textbf{0.415}&\textbf{0.316}
			&0.996&1.640&\textbf{0.920}&\textbf{1.337}&0.655&0.783&\textbf{0.551}&\textbf{0.599}
			&0.426&0.352&\textbf{0.418}&\textbf{0.342}\\
			&\textbf{+RevIN}&0.435&0.401&\textbf{0.424}&\textbf{0.381}
			&0.592&0.587&\textbf{0.519}&\textbf{0.460}
			&0.482&0.468&\textbf{0.447}&\textbf{0.416}
			&0.481&0.478&\textbf{0.438}&\textbf{0.386} \\
			&\textbf{+Dish-TS}&0.562&0.774&\textbf{0.493}&\textbf{0.468}
			&0.930&2.105&\textbf{0.704}&\textbf{1.003}
			&0.509&0.496&\textbf{0.476}&\textbf{0.395}&0.540&0.531&\textbf{0.502}&\textbf{0.476}
			\\
			&\textbf{+SAN}&0.483&0.521&\textbf{0.415}&\textbf{0.372}
			&0.457&0.410&\textbf{0.404}&\textbf{0.338}
			&0.496&0.499&\textbf{0.457}&\textbf{0.448}
			&0.418&\textbf{0.350}&\textbf{0.407}&0.356
			\\
			&\textbf{+FAN}&0.492&0.458&\textbf{0.482}&\textbf{0.457}
			&0.562&0.591&\textbf{0.533}&\textbf{0.528}&0.513&0.502&\textbf{0.486}&\textbf{0.448}&0.473&0.443&\textbf{0.462}&\textbf{0.419}\\ 
			\midrule
			\multirow{5}{*}{\rotatebox[origin=c]{90}{\textbf{Traffic}}} & \textbf{None} &0.288&0.554&\textbf{0.286}	&	\textbf{0.530}
			&0.425&0.830&\textbf{0.402}	&	\textbf{0.780}&0.594&0.986&\textbf{0.428}	&	\textbf{0.830}
			&0.298&\textbf{0.443}	&	\textbf{0.296}	&	\textbf{0.443}
			\\
			&\textbf{+RevIN}&0.284&0.421&\textbf{0.280}	&	\textbf{0.416}&0.457&0.874&\textbf{0.394}	&	\textbf{0.753}
			&\textbf{0.289}	&	\textbf{0.412}	&	0.293	&	0.415&\textbf{0.295}	&	\textbf{0.445}	&	\textbf{0.295}	&	\textbf{0.445} \\
			&\textbf{+Dish-TS}&0.293&0.438&\textbf{0.278}	&	\textbf{0.417}&0.445&0.861&\textbf{0.403}	&	\textbf{0.761}
			&\textbf{0.305}	&	\textbf{0.429}	&	0.309	&	0.430
			&0.314&0.461&\textbf{0.310}	&	\textbf{0.457}
			\\
			&\textbf{+SAN}&0.292&\textbf{0.432}&\textbf{0.291}	&	0.437&0.419&0.753&\textbf{0.394}	&	\textbf{0.716}&\textbf{0.294}	&	\textbf{0.427}	&	0.297	&	0.431
			&0.407&0.651&\textbf{0.363}	&	\textbf{0.569}\\
			&\textbf{+FAN}&0.316&0.454&\textbf{0.301}	&	\textbf{0.438}&0.315&0.457&\textbf{0.301}	&	\textbf{0.445}
			&0.340&0.468&\textbf{0.324}	&	\textbf{0.454}
			&0.302&0.432&\textbf{0.298}	&	\textbf{0.431}
			\\ 
			\midrule
			\multirow{5}{*}{\rotatebox[origin=c]{90}{\textbf{Weather}}} & \textbf{None} &0.280&0.229&\textbf{0.268}&\textbf{0.223}&0.396&0.401&\textbf{0.311}&\textbf{0.267}
			&0.302&0.256&\textbf{0.285}&\textbf{0.239}&0.307&\textbf{0.254}&\textbf{0.299}&0.260\\
			&\textbf{+RevIN}&\textbf{0.258}&\textbf{0.221}&0.259&0.224&0.294&0.279&\textbf{0.274}&\textbf{0.243}
			&\textbf{0.267}&\textbf{0.233}&0.269&\textbf{0.233}&0.283&\textbf{0.252}&\textbf{0.282}&\textbf{0.252}
			\\
			&\textbf{+Dish-TS}&0.277&\textbf{0.225}&\textbf{0.276}&0.226
			&0.316&0.296&\textbf{0.295}&\textbf{0.276}
			&0.296&0.258&\textbf{0.285}&\textbf{0.240}
			&0.301&\textbf{0.240}&\textbf{0.299}&\textbf{0.240}\\
			&\textbf{+SAN}&\textbf{0.277}&\textbf{0.225}&0.279&0.227
			&0.286&0.261&\textbf{0.276}&\textbf{0.245}
			&\textbf{0.271}&\textbf{0.228}&0.276&0.235&0.279&\textbf{0.226}&\textbf{0.277}&0.227
			\\
			&\textbf{+FAN}&\textbf{0.281}&\textbf{0.231}&0.283&0.234&0.285&0.246&\textbf{0.280}&\textbf{0.245}
			&0.288&0.236&\textbf{0.276}&\textbf{0.230}
			&\textbf{0.276}&\textbf{0.228}&0.278&0.229\\
			\midrule[1.2pt]
			\multicolumn{2}{c|}{\textbf{1$^{st}$ count}}&4&6&\textbf{26}&\textbf{24}&0&1&\textbf{30}&\textbf{29}&7&6&\textbf{24}&\textbf{25}&3&11&\textbf{28}&\textbf{23}
			\\ 
			
			\bottomrule[1.2pt]
		\end{tabular}
	}
	\caption{Results of the main experiment}
\end{table*}

\subsection{Training Strategy}
Although APT is conceptually simple, learning timestamp-conditioned affine parameters resembles self-supervised clustering, potentially forming a $\mathit{bi\mbox{-}level}$ optimization problem~\cite{gould2016differentiating}, similar to SAN~\cite{liu2023adaptive}.

To stabilize training and ensure effective learning of timestamp semantics, we first freeze the normalization and backbone, and train APT with additional epochs using the following self-supervised losses:

\subsubsection{Orthogonal Loss. }
To ensure diversity among timestamp and prototype embeddings, we encourage orthogonality within $\textbf{\textsl{E}} = \{\textbf{\textsl{T}}^{\mathrm{TiD}},\textbf{\textsl{T}}^{\mathrm{DiW}},\textbf{\textsl{P}}\}$. The loss is defined as:
\begin{equation}\label{orth}
	Loss_{orth} = ||\textbf{\textsl{E}}^\top \textbf{\textsl{E}} \odot (1 - \textbf{\textsl{I}})||^2_{2} + || \textbf{\textsl{I}} - \textbf{\textsl{E}}^\top \textbf{\textsl{E}}\odot \textbf{\textsl{I}} ||^2_{2}
\end{equation}
Where $\odot$ is the Hadamard product, $I$ is the identity matrix, and $||_{2}$ is the L2 norm. $Loss_{orth}$ penalizes correlation between embeddings and enforces unit-norm behavior, following Barlow Twins~\cite{zbontar2021barlow}.

\subsubsection{Load Balancing Loss. }
Prototype matching may lead to imbalanced usage, where few prototypes dominate. Inspired by MoE imbalance issues~\cite{wang2024auxiliary}. To mitigate this, we introduce a load balancing loss that encourages a uniform distribution over prototype usage. 
This method belongs to batch-wise loss, where we penalize the prototype weights $\textbf{\textsl{W}} \in \mathbb{R}^{B \times N}$ in each batch in Equation~\ref{w}:
\begin{equation}\label{balance}
	Loss_{balance} = \sum_{j=1}^{N} (\frac{\sum_{i=1}^{B} \textbf{\textsl{W}}_{i,j}}{\sum_{i=1}^{B}\sum_{j=1}^{N} \textbf{\textsl{W}}_{i,j}} - \frac{1}{N})^2
\end{equation}

\subsubsection{Affine Regularization Loss. }To prevent trivial or unstable affine parameters, we follow self-supervised representation regularization strategies~\cite{ermolov2021whitening, bardes2022vicreg}. Taking $\gamma$ as an example, the loss is:
\begin{equation}\label{r}
	Loss_{R} = \frac{1}{B}  (1- \sqrt{\sum_{i=1}^{B} \mathbf{\gamma}_i^2})^2 + \frac{1}{B} (\sum_{i=1}^{B} \mathbf{\gamma}_i)^2
\end{equation}

\subsubsection{Training Processing. }In the additional epochs of freezing the backbone and normalization strategy, the overall pretraining objective for APT is:
\begin{equation}\label{apt}
	Loss_{APT} = Loss_{orth} + Loss_{balance} + Loss_{R}
\end{equation}
Once pretrained, APT is jointly optimized with the normalization and forecasting modules under standard regression loss (e.g., MSE). If required by the normalization strategy (e.g., SAN), its loss can be included:
\begin{equation}\label{main}
	Loss_{main} = Loss_{normal} + Loss_{MSE}
\end{equation}

\begin{figure*}[t]
	\centering
	\includegraphics[width=0.8\textwidth]{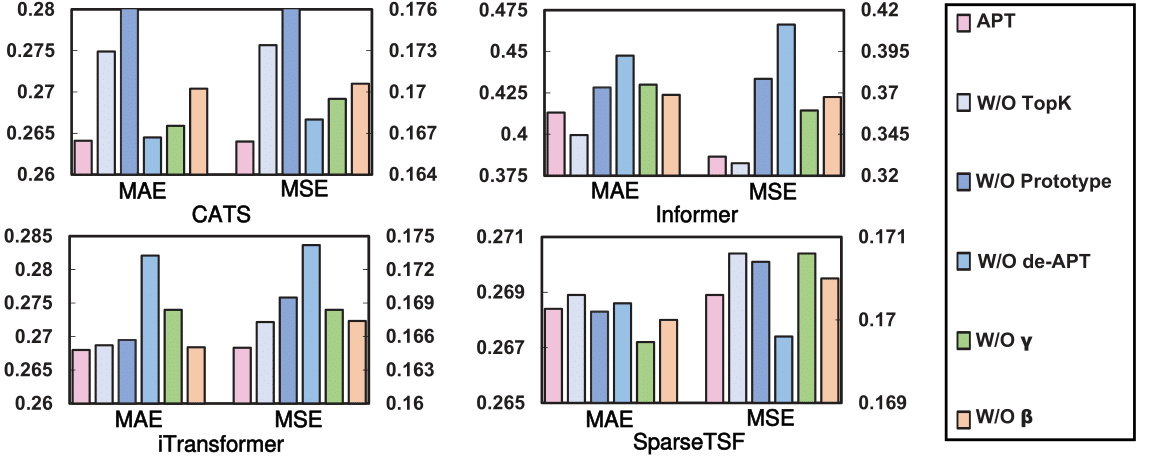
	}
	\caption{The ablation study results of APT components on the ECL dataset, $L=336, H=336$.}
	\label{xr1}
\end{figure*}

\section{Experiments}
\subsection{Experimental Setup}
\subsubsection{Baselines.} As a plug-in designed to mitigate distribution shift, APT is compatible with any time series forecasting backbone and normalization strategy. To validate its effectiveness, we select Informer~\cite{zhou2021informer}, iTransformer~\cite{liuitransformer}, SparseTSF~\cite{linsparsetsf}, and CATS~\cite{DBLP:conf/nips/Kim00K24} as the backbone, and RevIN~\cite{kim2021reversible}, Dish-TS~\cite{fan2023dish}, SAN~\cite{liu2023adaptive}, and FAN~\cite{DBLP:conf/nips/YeDZG24} as normalization strategies.

\subsubsection{Datasets \& Metrics.} We select six datasets for main experiments: ECL, ETTh1, ETTh2, Exchange, Traffic, Weather. The historical length $L$ is 336, the main experiments report average results over forecasting lengths $H = \{$96, 192, 336, 720$\}$, and the average value is reported, while extra experiments use $H = 336$. Metrics include mean squared error~(MSE) and mean absolute error~(MAE).

\subsubsection{Details.} We follow the BasicTS benchmark setup~\cite{shao2024exploring}. APT uses “Day in Week” and “Time in Day” as timestamp features, except on Exchange~(only “Day of Week”). The number of prototypes depends on sampling rate: 5 for daily, 30 for hourly, and 40 for 10-minute frequency. All embeddings are 20-dimensional with MLP hidden size 32, yielding 1.5K–5K parameters. APT trains with a learning rate of 5e-5 for 1–5 extra epochs until convergence. The optimizer is shared with other modules, with gradient updates controlled by loss weight $\lambda$. More details are in Appendix~\ref{baselines} and~\ref{Exp}. 

\subsection{Main Results}
On datasets with strong distribution shifts like ETTh1/2 and Exchange, APT yields 4\%–40\% performance gains across diverse backbone–normalization combinations. Combined with robust normalization, even the weaker Informer matches or surpasses state-of-the-art backbones.

CATS, iTransformer, and SparseTSF are among the most advanced backbones, while RevIN, FAN and SAN are the most effective normalization strategies. Even on stable datasets like ECL, where performance typically plateaus, APT still yields consistent gains and often achieves the best overall results with negligible risk of degradation. More and combined with data analysis are provided in Appendix~\ref{mainexp}.

\subsection{Ablation Study}
In Figure~\ref{xr1} and Appendix~\ref{addabl}, we conduct ablations from two perspectives: component-wise and training strategy.
\subsubsection{Component Ablation:}We evaluate APT by removing key components:

\begin{itemize}
	\item\textbf{W/O Top-$k$}: Remove Top-$k$ from Equation~\ref{softmax};
	\item \textbf{W/O Prototype}: Use raw timestamp embeddings without prototype matching;
	\item \textbf{W/O de-APT}: Remove inverse transformation of APT;
	\item \textbf{W/O $\gamma$ or $\beta$}: Remove scaling or bias in affine transformation;\end{itemize}

Removing \textbf{de-APT} degrades performance, as it forces the model to reconstruct distributional properties.
Top-$k$ is similar to expert activation and mitigates feature over-smoothing by enforcing sparsity during prototype aggregation. In contrast, prototype learning alleviates the challenges of few-shot learning by compressing timestamp semantics, improving robustness and preventing overfitting.
Between affine components, \textbf{$\gamma$} is more crucial than the bias $\beta$ under distribution shift, because mean drift will generate abundant supervisory signals to guide the backbone to adapt.

\subsubsection{Training Strategy Ablation:}APT learns timestamp-aware affine parameters via self-supervised pretraining, while freezing backbone and normalization. Table~\ref{abl} and Appendix~\ref{Exp} show the effect of ablating components:
\begin{itemize}
	\item\textbf{W/O $\lambda$}: Remove the loss weighting coefficient from APT pre-training, sharing learning rate with the backbone;
	\item \textbf{W/O $Loss_{orth}$ or $Loss_{balance}$ or $Loss_{R}$}: Remove orthogonality, load balancing, or affine regularization loss;
	\item \textbf{W/O $Loss_{APT}$}: Skip pretraining entirely, optimizing APT only with the main loss.
\end{itemize}

\begin{figure}[t]
	\includegraphics[width=0.46\textwidth]{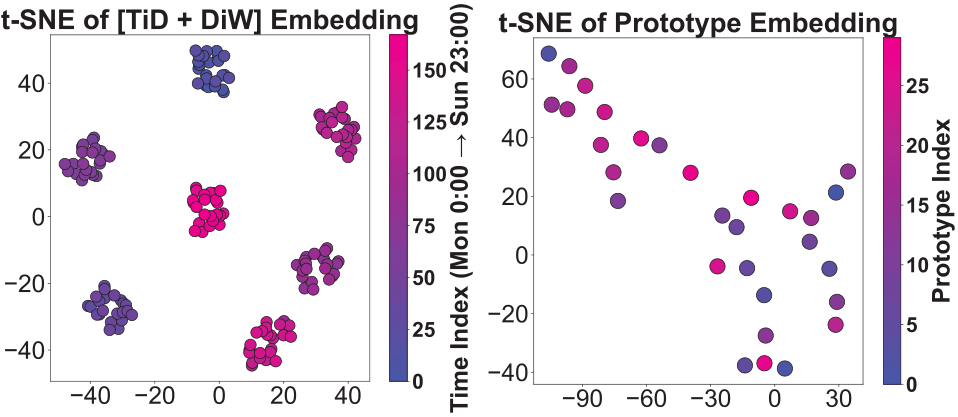}
	\caption{Visualization of APT timestamps and prototype embeddings on ECL dataset and iTransformer}
	\label{tsne}
\end{figure}

\begin{table*}[t]
	
	\centering 		 		 		
	\small
	\renewcommand\arraystretch{0.88}
	
	\setlength{\tabcolsep}{5pt}{
		\begin{tabular}{c|c|cc|cc|cc|cc|cc|cc}
			\toprule[1.2pt]
			\multicolumn{2}{c|}{\textbf{Methods}}  & \multicolumn{2}{c|}{\textbf{APT}} & \multicolumn{2}{c|}{\makecell[c]{\textbf{W/O} \\ $\lambda$}} &
			\multicolumn{2}{c|}{\makecell[c]{\textbf{W/O} \\ $Loss_{orth}$}} &
			\multicolumn{2}{c|}{\makecell[c]{\textbf{W/O} \\$Loss_{balance}$}} & \multicolumn{2}{c|}{\makecell[c]{\textbf{W/O} \\ $Loss_{R}$}} &
			\multicolumn{2}{c}{\makecell[c]{\textbf{W/O} \\ $Loss_{APT}$}}
			\\ 
			\midrule
			\multicolumn{2}{c|}{\textbf{Metrics}} & MAE & MSE & MAE & MSE& MAE & MSE& MAE & MSE& MAE & MSE& MAE & MSE\\
			\midrule[1.2pt]
			
			\multirow{4}{*}{\rotatebox[origin=c]{90}{\textbf{ECL}}} & \textbf{CATS} &\textbf{0.264}&\textbf{0.166}&0.269&0.169&\underline{0.265}&\underline{0.167}&0.266&0.169&\underline{0.265}&0.168&0.267&0.170\\
			&\textbf{Informer}&\underline{0.413}&\underline{0.332}&0.441&0.379&0.429&0.350&\textbf{0.412}&\textbf{0.329}&0.436&0.377&0.453&0.395 \\
			&\textbf{iTransformer}&\textbf{0.268}&\textbf{0.165}&\underline{0.269}&0.169&\textbf{0.268}&\textbf{0.165}&\underline{0.269}&\underline{0.167}&\underline{0.269}&0.168&0.270&0.168\\
			&\textbf{SparseTSF}&\textbf{0.268}&\textbf{0.170}&0.270&0.171&\underline{0.269}&0.171&\underline{0.269}&0.171&0.269&0.171&0.268&\underline{0.171}\\
			\midrule
			\multirow{4}{*}{\rotatebox[origin=c]{90}{\textbf{ETTh1}}} & \textbf{CATS} &\textbf{0.437}&\textbf{0.443}&0.451&0.458&\underline{0.438}&\underline{0.446}&0.448&0.460&0.439&\underline{0.446}&0.448&0.459\\
			&\textbf{Informer}&\underline{0.797}&\textbf{1.121}&0.856&1.249&0.839&1.215&1.009&1.495&\textbf{0.787}&\textbf{1.121}&0.818&\underline{1.198} \\
			&\textbf{iTransformer}&0.474&\underline{0.478}&\underline{0.472}&0.481&\underline{0.472}&0.479&\textbf{0.469}&\textbf{0.474}&0.487&0.498&0.483&0.491\\
			&\textbf{SparseTSF}&\textbf{0.438}&\textbf{0.451}&0.447&0.458&\underline{0.441}&\underline{0.453}&0.448&0.460&0.441&0.454&0.448&0.459\\
			\midrule
			\multirow{4}{*}{\rotatebox[origin=c]{90}{\textbf{Exchange}}} & \textbf{CATS} &\textbf{0.402}&\textbf{0.276}&0.411&0.286&0.407&0.290&\underline{0.406}&\underline{0.278}&0.402&\textbf{0.276}&0.560&0.500\\
			&\textbf{Informer}&\textbf{0.976}&\underline{1.490}&1.144&2.070&\underline{1.009}&1.789&\underline{1.009}&\textbf{1.506}&0.989&1.490&1.082&1.734\\
			&\textbf{iTransformer}&\textbf{0.502}&\textbf{0.445}&0.634&0.6549&0.629&0.669&0.559&0.531&\underline{0.544}&\underline{0.508}&0.655&0.785\\
			&\textbf{SparseTSF}&\textbf{0.431}&\textbf{0.316}&\underline{0.432}&\underline{0.321}&0.443&0.333&\textbf{0.431}&\textbf{0.316}&0.439&0.330&0.448&0.343\\
			\midrule
			\multirow{4}{*}{\rotatebox[origin=c]{90}{\textbf{Weather}}} & \textbf{CATS} &\textbf{0.283}&\textbf{0.242}&\underline{0.287}&\underline{0.243}&0.300&0.246&0.300&0.247&0.295&\underline{0.243}&0.301&0.246\\
			&\textbf{Informer}&\textbf{0.315}&\textbf{0.270}&\textbf{0.315}&\underline{0.271}&\underline{0.326}&0.284&0.328&0.291&0.410&0.389&0.554&0.816 \\
			&\textbf{iTransformer}&\textbf{0.293}&\textbf{0.251}&0.296&\underline{0.254}&\textbf{0.293}&\textbf{0.251}&\underline{0.295}&\underline{0.254}&0.296&\underline{0.254}&0.300&0.257\\
			&\textbf{SparseTSF}&\textbf{0.312}&\textbf{0.267}&\underline{0.314}&\textbf{0.267}&0.320&0.270&0.322&0.271&0.320&0.270&0.319&\underline{0.269}\\
			
			\bottomrule[1.2pt]
	\end{tabular}}
	\caption{Results of ablation studies on training strategies across four datasets. $L=336, H=336$}
	\label{abl}
\end{table*}

\begin{figure*}[t]
	\centering
	\includegraphics[width=0.95\textwidth]{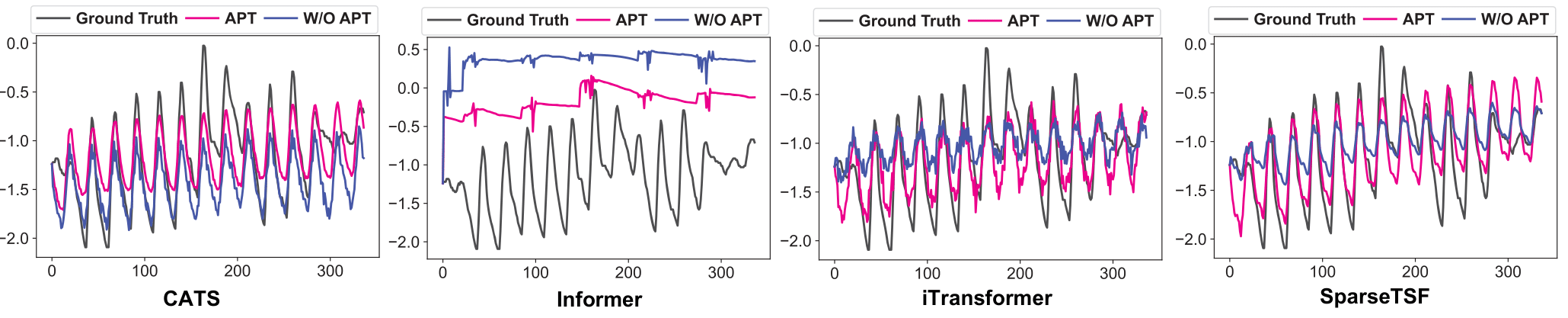}
	\caption{Visualization of forecasting results for different models on the ETTh2 dataset without normalization strategy}
	\label{caseetth21}
\end{figure*}
\begin{figure}[t]
	\centering
	\includegraphics[width=0.42\textwidth]{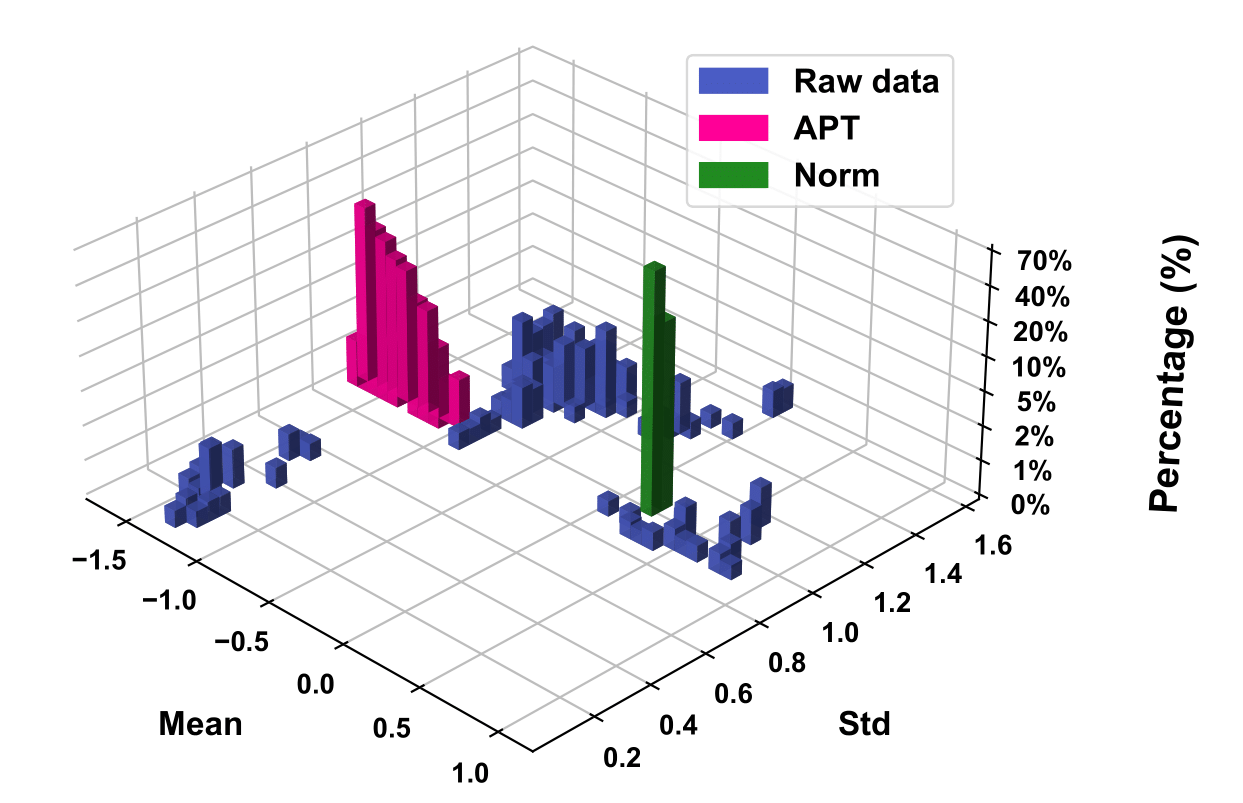}
	\caption{3D visualization of temporal distribution and ratio of forecasting pipeline at different stages on ETTh1.}
	\label{shift}
\end{figure}

Recommended learning rates for $\{$CATS, Informer, iTransformer, SparseTSF$\}$ are $\{$5e-3, 2e-4, 5e-4, 2e-3$\}$, with APT pretraining fixed at 5e-5. Accordingly, $\lambda$ is set to $\{$5e-2, 2.5e-1, 1e-1, 2.5e-2$\}$. Without $\lambda$,  APT shares the backbone’s rate, causing overfitting and unstable timestamp semantics.

Each pretraining loss targets a distinct goal: $ Loss_{orth}$ encourages embedding diversity, $Loss_{balance}$ prevents feature collapse~\cite{hua2021feature}, and $Loss_{R}$ stabilizes convergence for regression compatibility.
As they govern embeddings, prototype matching, and parameter scaling respectively, these losses are complementary and essential for APT’s stability. Omitting pretraining altogether~(W/O $Loss_{APT}$) causes APT to fail, with performance reverting to backbone levels.

\subsection{Visualization}
\textbf{t-SNE of Embedding:} Figure~\ref{tsne} shows the timestamp and prototype embeddings learned by APT with iTransformer on the ECL dataset. Timestamp embeddings form distinct clusters aligned with “Day in Week,” while “Time in Day” varies orthogonally in Appendix~\ref{visual}. Prototypes are also well-separated, but their embeddings still preserve certain timestamp-related structure as orthogonal losses serve as soft supervision.

\textbf{Forecasting cases:} Figure~\ref{caseetth21} presents forecasting cases on the 6th channel of ETTh2 without normalization. While most backbones, except Informer, capture periodic patterns well, they fail to handle distribution shifts over time. APT mitigates this issue by applying timestamp-conditioned affine transformations, independently of the backbone, to better align predictions with data distributions.

\textbf{Temporal distribution:} As shown in Figure~\ref{shift}, raw data exhibit severe distribution shift and its distributional features are disrupted, which is the phenomenon we emphasized earlier and it can cause great damage to backbone's learning of temporal patterns. Normalization maps series to zero mean and unit variance; while RevIN decouples distribution from patterns to improve robustness, it also suppresses informative distribution. In contrast, APT uses timestamps to encode global distribution features, forming compact yet discriminative representations adaptable to temporal shift.

Due to space limitations, the complete results and more experiments are provided in Appendix~\ref{Exp} to better understand APT:  data motivation~(\ref{datasets}), other plug-ins~(\ref{pluginstudy}), extra main results~(\ref{mainexp}), hyper-Parameter~(\ref{hyperstudy}), paramenter count~(\ref{paracount}), extra ablation study~(\ref{addabl}), cross-setting fine-tuning~(\ref{crosssetting}), visualization~(\ref{visual}) and discussion~(\ref{discussion}).

\section{Conclusion}
In this work, we propose Affine Prototype-Timestamp (APT), a lightweight and model-agnostic plug-in to enhance time series forecasting under distribution shifts. APT employs discretized timestamps and prototype learning to introduce timestamp-conditioned affine transformations, enabling forecasting models to recover global distributional features that may be suppressed by normalization or distorted by distribution shifts.

In future work, we will explore online adaptation for streaming data and incorporate external modalities such as text or images to enhance shift awareness, further strengthening the real-world applicability of deep forecasting models in domains like weather, finance, and energy.

\section*{Acknowledgment}
This work is supported by the NSFC underGrant Nos.
62372430 and 62502505, the Youth Innovation Promotion
Association CAS No.2023112, the Postdoctoral Fellowship
Program of CPSF under Grant Number GZC20251078, the
China Postdoctoral Science Foundation No.2025M771542 and
HUA Innovation fundings.

\bibliography{aaai2026}

\clearpage

\appendix

\section{Notation}

\begin{table}[h]
	\centering 		 		 		
	\renewcommand\arraystretch{0.9}
	\label{notation}
	\setlength{\tabcolsep}{1pt}
	{\begin{tabular}{c|c}
			\toprule[1.2pt]
			\textbf{Notation} & \textbf{Description}\\
			\midrule[1.2pt]
			$x$& Time series, especially historical parts \\
			\midrule
			$L$& Length of historical time series\\
			\midrule
			$y$& Time series, especially future parts\\
			\midrule
			$H$& Length of future time series\\
			\midrule
			$\boldsymbol{\mathcal{M}}$& Deep Time Series Forecasting Model\\
			\midrule
			$\boldsymbol{\mathcal{N}}$& Time Series Normalization Strategy\\
			\midrule
			$t$& Subscript, representing a certain moment in time\\
			\midrule
			$ts$& Timestamp information\\
			\midrule
			$\textbf{\textsl{T}}$& Timestamp embedding\\
			\midrule
			$\textbf{\textsl{P}}$& Prototype embedding, $\textbf{\textsl{p}} \in \textbf{\textsl{P}}$ .\\
			\midrule
			$\textbf{\textsl{S}}$& Similarity matrix\\
			\midrule
			$\textbf{\textsl{I}}$& Identity matrix\\
			\midrule
			$\textbf{\textsl{E}}$& General term for \textsl{T} and \textsl{P}\\
			\midrule
			$\textbf{\textsl{W}}$& Weights used for prototype weighting\\
			\midrule
			$\mu$& Mean\\
			\midrule
			$\sigma$& Variance\\
			\midrule
			$\gamma$& Affine parameters for scaling\\
			\midrule
			$\beta$& Affine parameters for bias\\
			\midrule
			\multirow{2}{*}{$\lambda$}&Learning rate factor for APT \\
			&typically (5e-5)/$lr$, $lr$ is learning rate of MSE\\ 
			
			\bottomrule[1.2pt]
	\end{tabular}}
	\caption{Explanation of notations leveraged in APT}
\end{table}

\section{Baselines}\label{baselines}
\subsection{Deep Time Series Forecasting Model}
\noindent\textbf{CATS}~\cite{DBLP:conf/nips/Kim00K24} is a time series forecasting model constructed entirely by cross-channel attention. It adopts the Patch strategy~\cite{nietime}, uses future-related parameters as queries and masks potentially interfering temporal information during forecasting. Inspired by the effectiveness of linear model forecasts~\cite{zeng2023transformers}, CATS strikes a balance between efficiency and modeling complex temporal dependencies.

\noindent\textbf{Informer}~\cite{zhou2021informer} is one of the pioneers in the field of deep time series forecasting. As a classic work, it reflects on the efficiency issues of Transformers in time series forecasting, proposes a sparse attention mechanism, and achieves effective prediction.

\noindent\textbf{iTransformer}~\cite{liuitransformer} addresses the limitations of previous work, which treated the time series dimension as a temporal token while ignoring inter-channel correlations, and the efficiency issues arising from increasing forecast lengths. To address these issues, it proposes treating different channels as tokens and applying an attention mechanism between channels, thereby achieving more effective forecasts than previous models.

\noindent\textbf{SparseTSF}~\cite{linsparsetsf} is a state-of-the-art linear forecasting model that further reduces the number of parameters. It divides time series into multiple periods through downsampling, and effectively forecasts time series through cross-period forecasting with parameter sharing and upsampling.

\subsubsection{Factors for selecting those backbones:} Informer is a pioneering work in long-term time series forecasting. As a classic forecasting model, it incorporates two temporal modeling strategies: Transformer and channel dependency. Even though its performance is limited in other works, we can prove that with the help of advanced normalization and APT, it can still achieve SOTA performance in many cases.

SparseTSF inherits from DLinear and offers better performance. It includes strategies such as MLP, channel independence, and patch. iTransformer is based on Transformer and has a channel interaction strategy that captures spatial information and temporal patterns better than Informer's channel dependency. Compared to iTransformer, CATS uses a more refined channel interaction method with cross-attention, and also has the patch strategy.
\subsection{Time Series Normalization Strategy}

\noindent\textbf{RevIN}~\cite{kim2021reversible} is a classic work that alleviates shift in time series forecasting. Time series are mutually coupled with patterns and distributions. Forecasting models mainly learn pattern information related to temporal dependencies, while distributions are considered components unrelated to forecasting. By separating distribution information before inputting it into the model through a normalization strategy and restoring that information after forecasting through inverse normalization, RevIN effectively improves the forecasting capabilities of various models.

\noindent\textbf{DishTS}~\cite{fan2023dish} believes that time series shift is constantly changing, not only within the time series, but also between the history and the future. To this end, it proposes using additional network learning distribution statistics and forecasting the future statistics required for inverse normalization.

\noindent\textbf{SAN}~\cite{liu2023adaptive} thinks that the scenarios explored by Dish-TS are still not comprehensive enough. The distribution of time series continues to shift, and the length of history and future has exceeded the scope of effective description. Therefore, it further compresses the learning of drift information to the patch level and attempts to learn the shift flow within a single sample.

\noindent\textbf{FAN}~\cite{DBLP:conf/nips/YeDZG24} actually switches the time series shift from a distribution perspective to time series decomposition. Non-stationary information is considered to be minor components that are difficult to learn effectively. Therefore, FAN uses adaptive filtering to input only information at the main frequency into the model, adopts the residuals from the additional network information, and adds residual forecasting terms after model forecasting.

\textbf{PS: }It should be noted that, with the exception of Informer, other forecasting models originally come with RevIN due to its irreplaceable performance gains, or do not have an affine transformation version. In the APT test, we first completely separated the normalization strategy from the model itself. When testing the addition of RevIN, we adopt the original version containing affine transformation.

\begin{table*}[h]
	\centering 		 		 		
	\small
	\renewcommand\arraystretch{0.9}
		\setlength{\tabcolsep}{3pt}{
			\begin{tabular}{c|c|c|c|c|c|c|c|c|c}
				\toprule[1.2pt]
				\textbf{Datasets} & \textbf{Domain} & \textbf{Period} & \textbf{Frequency} & \textbf{Channel}& \textbf{Split} & \textbf{Timestamps} & \textbf{Prototype} & \textbf{Top K} & \textbf{Param.}\\
				\midrule
				ECL & Energy & 2012 - 2014 & 1 hour & 321 & 7:1:2 & TiD\&DiW & 30 & 3 & 2.64K\\
				\midrule 
				ETTh1\&2 & Energy & 2016/7/1 0:00 - 2018/6/26 19:45 & 1 hour & 7 & 6:2:2 & TiD\&DiW & 30 & 3 & 2.64K\\
				\midrule 
				Exchange & Economy & 1990 - 2016 & 1 day & 8 & 7:1:2 & DiW & 5 & 2 & 1.82K\\
				\midrule 
				Traffic & Transportation & 2016/7/1 02:00 - 2018/7/2 01:00 & 1 hour & 862 & 7:1:2 & TiD\&DiW & 30 & 3 & 2.64K\\
				\midrule 
				Weather & Nature & 2020/1/1 0:10 - 2021/1/1 0:00 & 10 mins & 21 & 7:1:2 & TiD\&DiW & 40 & 4 & 5.1K\\
				\bottomrule[1.2pt]
		\end{tabular}}
		\caption{Dataset description and corresponding parameters}
		\label{datatable}
	\end{table*}
	
	\begin{table*}[h]
		\centering 		 		 		
		\renewcommand\arraystretch{0.9}
		\small
			\setlength{\tabcolsep}{3pt}{
				\begin{tabular}{c|c|c|c|c|c|c|c}
					\toprule[1.2pt]
					\textbf{Datasets} & \textbf{Overall missing rate} & \textbf{0 fill rate} & \textbf{Previous value fill rate} & \multicolumn{2}{c|}{\textbf{Min. rate and channel}}& \multicolumn{2}{c}{\textbf{Max. rate and channel}}\\
					\midrule
					ECL & 4.03\% & 1.08\% & 2.94\% &  0.10\% & 133  & 85.89\%&298 \\
					\midrule 
					ETTh1 & 7.38\% & 0.99\% & 6.38\%  & 3.65\% & 2& 14.44\%& 5 \\
					\midrule 
					ETTh2 & 23.68\% & 9.79\% & 13.89\%  & 4.24\%& 0 &72.68\%& 5 \\
					\midrule 
					Exchange & 8.60\% & 0\% & 8.60\%  & 4.38\%& 1 & 32.64\%&4\\
					\midrule 
					Traffic & 1.68\% & 0.89\% & 0.78\%  & 0.08\% & 267& 15.85\%&751\\
					\midrule 
					Weather & 3.39\% & 0.01\% & 3.38\%& 0.05\% & 17  & 10.44\%& 8\\
					\bottomrule[1.2pt]
			\end{tabular}}
			\caption{Detailed report on dataset missing rates}
			\label{missing}
		\end{table*}
		
		\section{Experiments}\label{Exp}
		\subsection{Experimental details}\label{details}
		Our experiments fully adopt the default configuration of the public-source and fair benchmark BasicTS\footnote{https://github.com/GestaltCogTeam/BasicTS}, including ADAM as the default optimizer, with each model having its own dedicated parameter configuration file for each dataset.
		All experiments are conducted on a single NVIDIA GeForce RTX 4090 GPU, with an Intel(R) Xeon(R) Gold 6338 CPU @ 2.00GHz, and each experiment is limited to 4 threads.
		
		\subsection{Dataset Information}\label{datasets}
		\textbf{ETTh1\&h2\footnote{https://github.com/zhouhaoyi/ETDataset}:} ETT is a series of datasets, where "1" and "2" represent different transformers, "h" represents the sampling rate per hour. APT does not adopt its 15-minute sampling rate dataset. The first six variables in each ETT data set are overload data, and the last one is oil temperature.
		
		\noindent\textbf{ECL\footnote{https://github.com/laiguokun/multivariate-time-series-data\label{data1}}:} ECL, i.e. the Electricity dataset, records the hourly electricity consumption of 321 clients between 2012 and 2014 (unit: kilowatt-hours).
		
		\noindent\textbf{Exchange\footref{data1}:} The ExchangeRate datasets includes daily exchange rates for eight countries: Australia, the United Kingdom, Canada, Switzerland, China, Japan, New Zealand, and Singapore.
		
		\noindent\textbf{Traffic\footnote{https://github.com/thuml/Autoformer\label{data2}}:} Trafﬁc is a collection of hourly data provided by the California Department of Transportation that describes road usage measured by various sensors on highways in the San Francisco Bay Area.
		
		\noindent\textbf{Weather\footref{data2}:} Weather records every 10 minutes throughout 2020 from Autoformer, including 21 meteorological indicators such as air temperature and humidity.
		
		Table~\ref{datatable} presents key information about the datasets, particularly the sampling rates. The primary parameters of APT are configured based on the sampling rate, including the timestamps we adopt, the number of prototypes, and top‑$\mathit{k}$. Furthermore, we show the parameter count of APT under the setting of channel-shared affine parameters, which is often significantly smaller than that of time series forecasting models. In experiments utilizing channel-wise affine parameters, the additional parameter count depends on the dataset's channel count and the embedding dimension~(fixed at 20), and the increase remains minimal.

		\subsubsection{Motivation verification from data perspective:} One of the motivations for APT is that local statistical normalization cannot address variations within time series, such as missing values or noise.
		
		\textbf{Noise} is affected by random factors from the sensor's own mechanisms and the external environment. As information coupled with each step of time series, noise is typically difficult to detect directly. We refer to BasicTS's discussion on model information for time series datasets, analyzing pattern stability and distribution drift, and conclude that ECL, Traffic, and Weather are low-noise datasets with stable patterns and low distribution drift, while ETT and ExchangeRate are the opposite.
		
		\textbf{Missing values} often occur in real-world scenarios due to sensor failures. Even benchmarks often contain a large amount of missing data, and prior data processing methods may vary. Due to the lack of more prior knowledge, we have broadly categorized missing value handling into two cases: zero-filling and previous-value-filling:
		
		\begin{itemize}
			\item Zero-filling: Directly fill missing signals with 0. Note that active filling in the Weather dataset is -9999, and cases where precipitation is 0 have been excluded.
			\item Previous-value-filling: When the current value cannot be obtained, the previous value is used. Since the probability of recording two completely identical floating-point numbers in succession is very low, we do not perform further exclusions, but otherwise we would misjudge small fluctuations in the values.
		\end{itemize}
		
		The summary of the missing rate is shown in Table~\ref{missing}. We have calculated the two types of missing rates for different datasets, as well as the maximum and minimum missing rates and their channels. The missing rate for ECL, Traffic, and Weather is one level lower than that of other datasets, and its main missing rate is often affected by only a few channels, with most channels remaining almost free of missing values.
		
		On the contrary, the ETT and Exchange datasets not only have fewer channels but also higher minimum missing rates, which implies that the statistical properties of each instance may be significantly affected. However, the impact of previous value filling on mean calculation is limited, but it has a profound detrimental effect on variance.
		
		In summary, the analysis of noise and missing rates indicates that the temporal information in ECL and similar datasets is more stable and reliable, while ETT and Exchange datasets are unreliable and exhibit severe distribution shift. These analyses will support certain phenomena in subsequent performance analysis and ablation studies of APT.

		\begin{table}[t]
			\centering 		 		 		
			\renewcommand\arraystretch{0.9}
				\setlength{\tabcolsep}{3pt}{
					\begin{tabular}{c|c|cc|cc|cc}
						\toprule[1.2pt]
						\multicolumn{2}{c|}{\textbf{Methods}}  & \multicolumn{2}{c|}{\textbf{APT}} & \multicolumn{2}{c|}{\textbf{FreTS}} & \multicolumn{2}{c}{\textbf{GLAFF}}\\
						\midrule
						\multicolumn{2}{c|}{\textbf{Metrics}} &MAE&MSE&MAE&MSE&MAE&MSE\\
						\midrule[1.2pt]
						\multirow{5}{*}{\rotatebox[origin=c]{90}{\textbf{ETTh1}}}
						& 96 & \textbf{0.410} & 0.399 & \textbf{0.410} & \textbf{0.388} & 0.414 & 0.406 \\
						&192 & \textbf{0.430} & \textbf{0.432} & 0.438 & 0.434 & 0.462 & 0.471 \\
						&336 & \textbf{0.448} & \textbf{0.457} & 0.459 & 0.460 & 0.479 & 0.495 \\
						&720 & \textbf{0.477} & \textbf{0.475} & 0.513 & 0.516 & 0.542 & 0.568 \\
						\cmidrule{2-8}
						&Avg. & \textbf{0.441} & \textbf{0.441} & 0.455 & 0.450 & 0.475 & 0.485 \\
						\midrule
						\multirow{5}{*}{\rotatebox[origin=c]{90}{\textbf{Weather}}}
						& 96 & 0.208 & \textbf{0.158} & \textbf{0.207} & 0.159 & 0.211 & 0.165 \\
						&192 & \textbf{0.248} & \textbf{0.201} & 0.249 & 0.204 & 0.250 & 0.220 \\
						&336 & \textbf{0.284} & \textbf{0.250} & 0.292 & 0.260 & 0.288 & 0.260 \\
						&720 & 0.355 & \textbf{0.323} & \textbf{0.347} & 0.334 & 0.350 & 0.359 \\\cmidrule{2-8}
						&Avg. & \textbf{0.269} & \textbf{0.233} & 0.274 & 0.239 & 0.275 & 0.251 \\
						\bottomrule[1.2pt]
				\end{tabular}}
				\caption{Performance comparison between APT and other plug-ins on iTransformer. $H$ = 336}
				\label{plugin}
			\end{table}
			
			\begin{table*}[t]
				\small
				\renewcommand\arraystretch{0.9}
				\centering 		 		 		
				\begin{tabular}{c|c|p{32pt}<{\centering}p{32pt}<{\centering}|p{32pt}<{\centering}p{32pt}<{\centering}|p{32pt}<{\centering}p{32pt}<{\centering}|p{32pt}<{\centering}p{32pt}<{\centering}}
					\toprule[1.2pt]
					\multicolumn{2}{c|}{\textbf{Methods}}  & \multicolumn{2}{c|}{\textbf{APT}} & 	\multicolumn{2}{c|}{\makecell[c]{+ \\ \textbf{Double prototype}}} &
					
					\multicolumn{2}{c|}{\makecell[c]{+ \\ \textbf{Identity embedding}}} & \multicolumn{2}{c}{\makecell[c]{+ \\ \textbf{“Day in Month”}}}
					\\ 
					\midrule
					\multicolumn{2}{c|}{\textbf{Metrics}} & MAE & MSE & MAE & MSE& MAE & MSE& MAE & MSE\\
					\midrule[1.2pt]
					
					\multirow{4}{*}{\rotatebox[origin=c]{90}{\textbf{ECL}}} & \textbf{CATS} &\textbf{0.264}&\textbf{0.166}&\underline{0.269}&\underline{0.170}&0.270&\underline{0.170}&0.273&0.191\\
					&\textbf{Informer}&\textbf{0.413}&\textbf{0.332}&\underline{0.419}&\underline{0.346}&0.512&0.613&0.443&0.389 \\
					&\textbf{iTransformer}&\textbf{0.268}&\textbf{0.165}&0.268&\underline{0.166}&\underline{0.270}&0.169&0.270&0.169\\
					&\textbf{SparseTSF}&\textbf{0.268}&\textbf{0.170}&\textbf{0.268}&\textbf{0.170}&\textbf{0.268}&\textbf{0.170}&\textbf{0.268}&\textbf{0.170}\\
					\midrule
					\multirow{4}{*}{\rotatebox[origin=c]{90}{\textbf{ETTh1}}} & \textbf{CATS} &\underline{0.437}&\underline{0.443}&\textbf{0.436}&\textbf{0.442}&0.447&0.461&0.450&0.463\\
					&\textbf{Informer}&0.797&\textbf{1.121}&\textbf{0.790}&\underline{1.154}&\underline{0.792}&1.293&0.951&1.967 \\
					&\textbf{iTransformer}&\underline{0.474}&\underline{0.478}&\textbf{0.470}&\textbf{0.477}&\underline{0.474}&0.482&0.526&0.599\\
					&\textbf{SparseTSF}&\underline{0.438}&\underline{0.451}&0.443&0.455&\textbf{0.437}&\textbf{0.450}&0.442&0.454\\
					\midrule
					\multirow{4}{*}{\rotatebox[origin=c]{90}{\textbf{Exchange}}} & \textbf{CATS} &\underline{0.402}&\underline{0.276}&0.406&0.284&\textbf{0.397}&\textbf{0.265}&0.410&0.294\\
					&\textbf{Informer}&0.976&1.490&\underline{0.970}&\textbf{1.380}&\textbf{0.947}&\underline{1.410}&1.120&1.884\\
					&\textbf{iTransformer}&\underline{0.502}&\underline{0.445}&0.558&0.540&\textbf{0.477}&\textbf{0.396}&0.652&0.743\\
					&\textbf{SparseTSF}&\textbf{0.431}&\textbf{0.316}&\underline{0.442}&\underline{0.333}&0.447&0.326&0.459&0.358\\
					\midrule
					\multirow{4}{*}{\rotatebox[origin=c]{90}{\textbf{Weather}}} & \textbf{CATS} &\textbf{0.283}&\textbf{0.242}&0.308&0.251&\underline{0.293}&\underline{0.246}&0.304&0.259\\
					&\textbf{Informer}&\textbf{0.315}&\underline{0.270}&\underline{0.323}&\textbf{0.264}&0.364&0.346&0.333&0.702 \\
					&\textbf{iTransformer}&\textbf{0.293}&\textbf{0.251}&\underline{0.297}&\underline{0.256}&0.303&\underline{0.256}&\underline{0.297}&0.257\\
					&\textbf{SparseTSF}&\textbf{0.312}&\textbf{0.267}&\underline{0.316}&0.270&0.317&0.269&0.317&\underline{0.268}\\
					
					\bottomrule[1.2pt]
				\end{tabular}
				\caption{Results of hyper-parameter sensitivity study  across four datasets. $L=336, H=336$}
				\label{hyper}
			\end{table*}
			
			\subsection{Plug-in Supplementary Study}~\label{pluginstudy}
			In this section, we added two plug-ins unrelated to distribution drift for supplementary study:
			\begin{itemize}
				\item \textbf{FreDF}~\cite{wang2025fredf}: FreDF is a work that explores the autocorrelation of time series from a frequency domain perspective, using an additional frequency domain loss function to enhance the modeling capabilities of time series correlation.
				\item \textbf{GLAFF}~\cite{wang2024rethinking}: GLAFF is a classic study on the role of timestamp information in time series forecasting. It introduces global information to the output of the forecasting pipeline through additional timestamp modeling. Its motivation is completely different from that of APT. GLAFF focuses on modeling global pattern information and weighting it with the forecasting results, while APT improves the normalization and denormalization processes through affine transformations to mitigate time series distribution shift, and the dynamic affine parameters provided by APT for each instance have no regression capability.
			\end{itemize}
			As shown in Table~\ref{plugin}. APT outperforms both FreDF and GLAFF by 2\%–10\% on ETTh1 and Weather when applied to iTransformer. Moreover, APT achieves better performance with longer forecasting horizons $H$, implying its stronger global temporal awareness that support long-term dependency modeling. In general, longer prediction horizons exacerbate distribution shift in time series, and this trend aligns well with both the motivation and design of APT timestamps serve better as a tool to alleviate distribution shift than as a pattern modeling signal.
			
			APT’s consistent advantage over GLAFF further supports this view. In our experiments, iTransformer alone contains 6.6M parameters, while GLAFF, as a plug-in, introduces an additional 1.8M parameters and requires twice the training time compared to APT. In contrast, APT only introduces 2.64K to 5.1K additional parameters and incurs almost no increase in training time. We attribute this inefficiency to GLAFF’s reliance on complex attention mechanisms to capture global timestamp information. Its inferior performance and higher cost inevitably limit its practical applicability and hinder deeper exploration of timestamp-driven time series modeling.
			
			\subsection{Main Experiment}\label{mainexp}
			The complete results of the main experiment are shown in Tables~\ref{main1} and ~\ref{main2}.
			
			APT yields the most significant improvements on Informer, achieving 3\%–50\% gains across most datasets. It consistently enhances Informer's performance when combined with any normalization strategy. Specifically, on ECL and Traffic, combining Informer + APT with FAN achieves performance comparable to or exceeding that of other backbones. On Weather, APT reaches similar results regardless of the normalization strategy used.
			We consider that this result may affect the role of channel dependence, an early strategy in multivariate time series forecasting. In datasets with stable time series patterns, channel dependence does not show a significant performance gap compared to other methods such as channel independence and interaction, but this still requires further research.
			
			On ETT and Exchange, APT improves performance by 2\%–10\% across various backbone–normalization combinations, with virtually no cases of degradation. These datasets are known for exhibiting significant distribution shift and containing missing values, which often impair the effectiveness of statistical normalization methods. By leveraging timestamps, APT introduces global temporal distribution awareness and dynamically generates affine parameters, effectively adapting to time series distribution shift.
			
			On more stable datasets such as ECL, Traffic, and Weather, the improvements from APT are more modest. Without other normalization strategies, APT brings 2\%–10\% gains to the backbone; when combined with normalization, the gains are typically limited to 1\%–5\%, and in some cases, APT may not yield further benefits.
			Given that models like CATS already achieve strong performance on these datasets—and that additional normalization often brings little improvement—APT performs reasonably within expectations. Moreover, APT is lightweight, adds negligible training overhead, and frequently enables state-of-the-art performance. Thus, incorporating APT can be considered non-intrusive and low-risk.
			
			In summary, APT consistently enhances forecasting performance with minimal overhead. On challenging datasets such as ETT and Exchange, which exhibit significant distribution shift and data quality issues, APT achieves SOTA results with negligible risk of degradation. On more stable datasets like ECL, Traffic, and Weather, where strong backbones and normalization strategies already saturate performance, APT remains non-intrusive and yields moderate gains. Its lightweight design and broad compatibility make APT a reliable and effective enhancement to modern time series forecasting pipelines.
			
			\begin{table*}[h]
				\centering 		 		 		
				\small
					\setlength{\tabcolsep}{3pt}{
						\begin{tabular}{c|p{50pt}<{\centering}|p{50pt}<{\centering}|p{50pt}<{\centering}|p{50pt}<{\centering}||p{35pt}<{\centering}|p{35pt}<{\centering}|p{35pt}<{\centering}|p{35pt}<{\centering}||p{35pt}<{\centering}}
							\toprule[1.2pt]
							\multirow{2}{*}{\textbf{Module}} & \multicolumn{4}{c||}{\textbf{Backbone}}& \multicolumn{4}{c||}{\textbf{Normalization}} & \multirow{2}{*}{\textbf{APT}}\\
							\cmidrule{2-9}
							& \textbf{CATS} & \textbf{Informer} & \textbf{iTransformer} & \textbf{SparseTSF}& \textbf{RevIN} & \textbf{Dish-TS} & \textbf{SAN} & \textbf{FAN} & \\
							\midrule
							\textbf{96} & 1.401M & \multirow{4}{*}{11.329M}& 6.528M &  0.081K&\multirow{4}{*}{0.014K}& \multirow{4}{*}{4.718k}&0.369M&85.294K&\multirow{4}{*}{2.54K} \\
							
							\textbf{192} & 1.401M & & 6.577M  & 0.137K&&&0.377M &97.678K&   \\
							\textbf{336} & 1.401M & & 6.651M  & 0.221K&& &0.389M &0.116M& \\
							\textbf{720} & 1.402M & & 6.848M  & 0.445K&&&0.422M&0.166M&  \\
							\bottomrule[1.2pt]
					\end{tabular}}
					\caption{Parameter count reported for the backbone and normalization on the ETTh1 dataset. M = $1e^6$, K = $1e^3$}
					\label{para}
				\end{table*}
				
				\begin{table}[t]
					\centering 		 		 		
					\small
						\setlength{\tabcolsep}{3pt}{
							\begin{tabular}{c|c|c|c|c|c}
								\toprule[1.2pt]
								\multicolumn{2}{c|}{\textbf{Embedding Size}}  & \textbf{10} & \textbf{20$^{*}$} & \textbf{30} & \textbf{50}\\
								\midrule
								\multicolumn{2}{c|}{\textbf{Metrics}} &MSE&MSE&MSE&MSE\\
								\midrule[1.2pt]
								\multirow{4}{*}{\rotatebox[origin=c]{90}{\textbf{ECL}}}
								&CATS & 0.167 & \textbf{0.166} & \textbf{0.166} & 0.167\\
								&Informer & 0.344 & 0.331 & 0.360 & \textbf{0.327}\\
								&iTransformer & 0.167 & \textbf{0.165} & 0.168 & 0.166\\
								&SparseTSF & \textbf{0.170} & \textbf{0.170} & \textbf{0.170}& \textbf{0.170}\\
								\midrule
								\multirow{4}{*}{\rotatebox[origin=c]{90}{\textbf{ETTh1}}}
								&CATS & \textbf{0.442} & 0.443 & 0.443 & 0.444\\
								&Informer & 1.178 & 1.195 & 1.197 & \textbf{1.168}\\
								&iTransformer & 0.486 & \textbf{0.478} & 0.480 & 0.489\\
								&SparseTSF & 0.452 & \textbf{0.451} & 0.453 & 0.454\\
								\midrule
								\multirow{4}{*}{\rotatebox[origin=c]{90}{\textbf{Exchange}}}
								&CATS & 0.286 & 0.285 & 0.325 & \textbf{0.275}\\
								&Informer & 1.528 & 1.510 & \textbf{1.356} & 1.750\\
								&iTransformer & 0.452 & \textbf{0.444} & 0.460 & 0.519\\
								&SparseTSF & 0.321 & \textbf{0.316} & 0.340 & 0.348\\
								\midrule
								\multirow{4}{*}{\rotatebox[origin=c]{90}{\textbf{Weather}}}
								&CATS & 0.247 & 0.245 & 0.244 & \textbf{0.242}\\
								&Informer & 0.286 & \textbf{0.270} & 0.280& 0.281\\
								&iTransformer & 0.253 & \textbf{0.252} & 0.255 & 0.253\\
								&SparseTSF & \textbf{0.266} & \textbf{0.266} & 0.268 & 0.268\\
								\midrule
								
								\bottomrule[1.2pt]
						\end{tabular}}
						\caption{Results of embedding sizes on different datasets and backbones. $L$ = 336, $H$ = 336}
						\label{embeddingsize}
					\end{table}
					
					\begin{table*}[t]
						\centering

						\small
							
							\begin{tabularx}{0.8\textwidth}{c|*{7}{>{\centering\arraybackslash}X|}>{\centering\arraybackslash}X}
								\toprule[1.2pt]
								\multicolumn{9}{c}{Cross-\textbf{Backbone} $\boldsymbol{\mathcal{M}}$, $\mathcal{N}$ = None, Output length $H$ = 336}\\
								\midrule
								\multicolumn{1}{c|}{Source $\rightarrow$}  & \multicolumn{2}{c|}{\textbf{CATS}} & \multicolumn{2}{c|}{\textbf{Informer}} & \multicolumn{2}{c|}{\textbf{iTransformer}} & \multicolumn{2}{c}{\textbf{SparseTSF}}\\ 
								\midrule
								Target $\downarrow$& MAE & MSE & MAE & MSE & MAE & MSE & MAE & MSE\\
								\midrule[1.2pt]
								\textbf{CATS}& \textbf{0.436} & \textbf{0.443} & 0.439 & 0.449 & 0.438 & 0.444 & \textbf{0.436} & 0.446\\
								\midrule
								\textbf{Informer}& 0.817 & \textbf{1.094} & \textbf{0.797} & 1.195 & 0.805 & 1.134 & 0.804 & 1.158\\
								\midrule
								\textbf{iTransformer}& \textbf{0.465} & \textbf{0.470} & 0.472 & 0.478 & 0.473 & 0.478 & 0.472 & 0.478\\
								\midrule
								\textbf{SparseTSF}& \textbf{0.438} & \textbf{0.451} & 0.443 & 0.455 & 0.442& 0.455 & \textbf{0.438} & \textbf{0.451}\\
							\end{tabularx}
							\begin{tabularx}{0.8\textwidth}{c|*{9}{>{\centering\arraybackslash}X|}>{\centering\arraybackslash}X}
								\toprule[1.2pt]
								\multicolumn{11}{c}{Cross-\textbf{Normalization} $\boldsymbol{\mathcal{N}}$, $\mathcal{M}$ = iTransformer, Output length $H$ = 336}\\
								\midrule
								\multicolumn{1}{c|}{Source $\rightarrow$}& \multicolumn{2}{c|}{\textbf{None}} & \multicolumn{2}{c|}{\textbf{RevIN}} & \multicolumn{2}{c|}{\textbf{Dish-TS}} & \multicolumn{2}{c|}{\textbf{SAN}} & \multicolumn{2}{c}{\textbf{FAN}}\\ 
								\midrule
								Target $\downarrow$& MAE & MSE & MAE & MSE & MAE & MSE & MAE & MSE& MAE & MSE\\
								\midrule[1.2pt]
								\textbf{None}& \textbf{0.473} & \textbf{0.478} & 0.489 & 0.496 & 0.482 & 0.487 & 0.485 & 0.492& 0.475 & \textbf{0.478}\\
								\midrule
								\textbf{RevIN}& 0.454 & 0.468 & 0.448 & \textbf{0.457} & 0.452 & 0.460 & 0.459 & 0.476& \textbf{0.447} & 0.458\\
								\midrule
								\textbf{Dish-TS}& 0.502 & \textbf{0.506} & 0.503 & 0.507 & 0.508 & 0.513 & 0.501 & \textbf{0.506}& \textbf{0.501} & \textbf{0.506}\\
								\midrule
								\textbf{SAN}&0.469 & \textbf{0.479} & \textbf{0.461} & 0.486 & 0.469 & 0.491 & 0.462 & 0.492& 0.469 & 0.490\\
								\midrule
								\textbf{FAN}&0.477 & 0.487 & \textbf{0.476} & \textbf{0.486} & 0.478 & 0.487 & 0.477 & 0.487& 0.479 & 0.489\\
							\end{tabularx}
							\begin{tabularx}{0.8\textwidth}{c|*{7}{>{\centering\arraybackslash}X|}>{\centering\arraybackslash}X}
								\toprule[1.2pt]
								\multicolumn{9}{c}{Cross-\textbf{Output length $H$ = 336}, $\mathcal{M}$ = iTransformer, $\mathcal{N}$ = None}\\
								\midrule
								\multicolumn{1}{c|}{Source $\rightarrow$}  & \multicolumn{2}{c|}{\textbf{96}} & \multicolumn{2}{c|}{\textbf{192}} & \multicolumn{2}{c|}{\textbf{336}} & \multicolumn{2}{c}{\textbf{720}}\\ 
								\midrule
								Target $\downarrow$& MAE & MSE & MAE & MSE & MAE & MSE & MAE & MSE\\
								\midrule[1.2pt]
								\textbf{96}&\textbf{0.410} & \textbf{0.389} & 0.411 & 0.398 & 0.430 & 0.416 & 0.440 & 0.415\\
								\midrule
								\textbf{192}&0.442 & 0.438 & \textbf{0.441} & \textbf{0.436} & 0.447 & 0.445 & 0.469 & 0.482\\
								\midrule
								\textbf{336}&0.477 & 0.491 & 0.504 & 0.524 & \textbf{0.473} & \textbf{0.478} & 0.514 & 0.531\\
								\midrule
								\textbf{720}&0.546 & 0.558 & 0.568 & 0.571 & 0.531 & 0.538 & \textbf{0.528} & \textbf{0.526}\\
								\bottomrule[1.2pt]
							\end{tabularx}
							\caption{Performance report of APT under one epoch of fine-tuning across Settings}
							\label{cross}
						\end{table*}

						\begin{table*}[t]
							\centering 		 		 		
							
							\small
							\renewcommand\arraystretch{0.75}
								\setlength{\tabcolsep}{4pt}{
									\begin{tabular}{c|c|c|cc|cc|cc|cc|cc|cc|cc|cc}
										\toprule[1.2pt]
										\multicolumn{3}{c|}{\textbf{Methods}}  & \multicolumn{4}{c|}{\textbf{CATS}} & \multicolumn{4}{c|}{\textbf{Informer}} & \multicolumn{4}{c|}{\textbf{iTransformer}} & \multicolumn{4}{c}{\textbf{SparseTSF}}\\ 
										\midrule
										\multicolumn{3}{c|}{\textbf{Affine}} & && \multicolumn{2}{c|}{\textbf{+APT}}&&& \multicolumn{2}{c|}{\textbf{+APT}}&&& \multicolumn{2}{c|}{\textbf{+APT}}&&& \multicolumn{2}{c}{\textbf{+APT}}\\
										\midrule
										\multicolumn{3}{c|}{\textbf{Metrics}} & MAE & MSE & MAE & MSE& MAE & MSE& MAE & MSE& MAE & MSE& MAE & MSE& MAE & MSE& MAE & MSE\\
										\midrule[1.2pt]
										\multirow{20}{*}{\rotatebox[origin=c]{90}{\textbf{ECL}}}
										& \multirow{4}{*}{\rotatebox[origin=c]{90}{\textbf{Nones}}} & 96 & 0.235 & 0.140 & \textbf{0.234} & \textbf{0.138} & 0.383 & 0.308 & \textbf{0.374} & \textbf{0.281} & \textbf{0.236} & \textbf{0.138} & 0.240 & 0.139 & \textbf{0.245} & \textbf{0.150} & \textbf{0.245} & \textbf{0.150} \\
										&  & 192 & 0.257 & 0.161 & \textbf{0.249} & \textbf{0.154} & 0.425 & 0.361 & \textbf{0.406} & \textbf{0.330} & 0.262 & 0.160 & \textbf{0.253} & \textbf{0.153} & 0.258 & 0.162 & \textbf{0.254} & \textbf{0.159} \\
										&  & 336 & 0.265 & 0.168 & \textbf{0.264} & \textbf{0.166} & 0.446 & 0.398 & \textbf{0.413} & \textbf{0.331} & 0.280 & 0.174 & \textbf{0.268} & \textbf{0.165} & 0.269 & 0.171 & \textbf{0.268} & \textbf{0.170} \\
										&  & 720 & \textbf{0.298} & \textbf{0.203} & 0.302 & 0.205 & 0.449 & 0.411 & \textbf{0.438} & \textbf{0.384} & 0.302 & 0.205 & \textbf{0.301} & \textbf{0.203} & 0.303 & 0.207 & \textbf{0.301} & \textbf{0.206} \\ \cmidrule{2-19}
										& \multirow{4}{*}{\rotatebox[origin=c]{90}{\textbf{RevIN}}} & 96 & \textbf{0.232} & \textbf{0.137} & 0.233 & 0.138 & 0.308 & 0.211 & \textbf{0.289} & \textbf{0.189} & \textbf{0.230} & \textbf{0.135} & 0.233 & 0.137 & 0.244 & 0.150 & \textbf{0.243} & \textbf{0.149} \\
										&  & 192 & 0.250 & \textbf{0.154} & \textbf{0.249} & \textbf{0.154} & 0.318 & 0.222 & \textbf{0.314} & \textbf{0.217} & 0.252 & 0.160 & \textbf{0.250} & \textbf{0.154} & 0.252 & 0.159 & \textbf{0.251} & \textbf{0.158} \\
										&  & 336 & \textbf{0.260} & \textbf{0.165} & \textbf{0.260} & \textbf{0.165} & 0.307 & 0.209 & \textbf{0.306} & \textbf{0.206} & 0.264 & 0.169 & \textbf{0.261} & \textbf{0.164} & 0.266 & 0.172 & \textbf{0.265} & \textbf{0.171} \\
										&  & 720 & \textbf{0.291} & \textbf{0.203} & 0.295 & 0.207 & 0.416 & 0.369 & \textbf{0.393} & \textbf{0.322} & \textbf{0.289} & \textbf{0.194} & 0.292 & \textbf{0.194} & \textbf{0.297} & \textbf{0.212} & \textbf{0.297} & \textbf{0.212} \\ \cmidrule{2-19}
										& \multirow{4}{*}{\rotatebox[origin=c]{90}{\textbf{Dish-ts}}} & 96 & 0.246 & 0.145 & \textbf{0.242} & \textbf{0.144} & 0.348 & 0.258 & \textbf{0.336} & \textbf{0.249} & \textbf{0.236} & \textbf{0.136} & \textbf{0.236} & \textbf{0.136} & 0.243 & 0.146 & \textbf{0.242} & \textbf{0.145} \\
										&  & 192 & \textbf{0.259} & \textbf{0.160} & \textbf{0.259} & \textbf{0.160} & 0.388 & 0.309 & \textbf{0.335} & \textbf{0.246} & 0.262 & 0.163 & \textbf{0.253} & \textbf{0.153} & \textbf{0.254} & \textbf{0.159} & \textbf{0.254} & \textbf{0.159} \\
										&  & 336 & 0.276 & 0.175 & \textbf{0.273} & \textbf{0.172} & \textbf{0.338} & \textbf{0.249} & 0.379 & 0.296 & 0.270 & 0.168 & \textbf{0.262} & \textbf{0.160} & 0.271 & 0.174 & \textbf{0.269} & \textbf{0.172} \\
										&  & 720 & \textbf{0.307} & \textbf{0.211} & \textbf{0.307} & 0.212 & 0.440 & 0.403 & \textbf{0.434} & \textbf{0.395} & \textbf{0.294} & 0.199 & \textbf{0.294} & \textbf{0.194} & \textbf{0.303} & \textbf{0.207} & \textbf{0.303} & 0.209 \\ \cmidrule{2-19}
										& \multirow{4}{*}{\rotatebox[origin=c]{90}{\textbf{SAN}}} & 96 & 0.243 & 0.141 & \textbf{0.241} & \textbf{0.140} & 0.290 & 0.186 & \textbf{0.264} & \textbf{0.160} & \textbf{0.233} & \textbf{0.134} & 0.236 & 0.136 & 0.240 & \textbf{0.139} & \textbf{0.239} & \textbf{0.139} \\
										&  & 192 & \textbf{0.266} & \textbf{0.165} & 0.267 & 0.166 & 0.301 & 0.199 & \textbf{0.296} & \textbf{0.196} & \textbf{0.248} & \textbf{0.148} & 0.250 & 0.151 & 0.255 & 0.154 & \textbf{0.253} & \textbf{0.153} \\
										&  & 336 & 0.281 & 0.179 & \textbf{0.280} & \textbf{0.178} & 0.352 & 0.259 & \textbf{0.313} & \textbf{0.208} & \textbf{0.262} & \textbf{0.161} & 0.264 & \textbf{0.161} & 0.270 & 0.167 & \textbf{0.269} & \textbf{0.166} \\
										&  & 720 & 0.316 & 0.217 & \textbf{0.315} & \textbf{0.216} & 0.454 & 0.420 & \textbf{0.422} & \textbf{0.362} & \textbf{0.291} & \textbf{0.192} & 0.296 & 0.193 & 0.319 & 0.226 & \textbf{0.305} & \textbf{0.205} \\\cmidrule{2-19}
										& \multirow{4}{*}{\rotatebox[origin=c]{90}{\textbf{FAN}}} & 96 & 0.238 & 0.140 & \textbf{0.234} & \textbf{0.137} & 0.243 & 0.144 & \textbf{0.237} & \textbf{0.141} & 0.248 & 0.147 & \textbf{0.240} & \textbf{0.141} & 0.236 & 0.138 & \textbf{0.233} & \textbf{0.136} \\
										&  & 192 & 0.251 & 0.154 & \textbf{0.249} & \textbf{0.152} & \textbf{0.251} & \textbf{0.152} & 0.252 & 0.153 & 0.259 & 0.159 & \textbf{0.254} & \textbf{0.156} & 0.250 & 0.153 & \textbf{0.248} & \textbf{0.152} \\
										&  & 336 & \textbf{0.267} & 0.168 & \textbf{0.267} & \textbf{0.167} & \textbf{0.266} & \textbf{0.163} & 0.271 & 0.171 & 0.278 & 0.175 & \textbf{0.271} & \textbf{0.170} & 0.266 & 0.167 & \textbf{0.265} & \textbf{0.165} \\
										&  & 720 & 0.305 & 0.206 & \textbf{0.300} & \textbf{0.205} & 0.303 & 0.200 & \textbf{0.296} & \textbf{0.199} & 0.311 & \textbf{0.209} & \textbf{0.309} & 0.210 & 0.301 & 0.204 & \textbf{0.300} & \textbf{0.203} \\
										\midrule[1.2pt]
										
										\multirow{20}{*}{\rotatebox[origin=c]{90}{\textbf{ETTh1}}}
										& \multirow{4}{*}{\rotatebox[origin=c]{90}{\textbf{Nones}}} & 96 & 0.416 & 0.404 & \textbf{0.400} & \textbf{0.382} & 0.865 & 1.119 & \textbf{0.669} & \textbf{0.896} & 0.450 & 0.430 & \textbf{0.410} & \textbf{0.389} & 0.392 & \textbf{0.374} & \textbf{0.390} & \textbf{0.374} \\
										&  & 192 & 0.451 & 0.460 & \textbf{0.422} & \textbf{0.420} & 0.898 & 1.263 & \textbf{0.736} & \textbf{1.018} & 0.479 & 0.481 & \textbf{0.441} & \textbf{0.436} & 0.424 & \textbf{0.420} & \textbf{0.422} & \textbf{0.420} \\
										&  & 336 & 0.468 & 0.490 & \textbf{0.436} & \textbf{0.443} & 0.819 & \textbf{1.153} & \textbf{0.797} & 1.195 & 0.497 & 0.512 & \textbf{0.473} & \textbf{0.478} & 0.446 & 0.457 & \textbf{0.438} & \textbf{0.451} \\
										&  & 720 & 0.541 & 0.591 & \textbf{0.484} & \textbf{0.481} & 0.982 & 1.487 & \textbf{0.930} & \textbf{1.427} & 0.568 & 0.591 & \textbf{0.528} & \textbf{0.526} & \textbf{0.496} & \textbf{0.493} & 0.497 & 0.497 \\ \cmidrule{2-19}
										& \multirow{4}{*}{\rotatebox[origin=c]{90}{\textbf{RevIN}}} & 96 & 0.411 & 0.396 & \textbf{0.400} & \textbf{0.382} & 0.567 & 0.640 & \textbf{0.441} & \textbf{0.442} & 0.420 & 0.410 & \textbf{0.410} & \textbf{0.399} & 0.391 & \textbf{0.374} & \textbf{0.390} & \textbf{0.374} \\
										&  & 192 & 0.432 & 0.437 & \textbf{0.421} & \textbf{0.420} & 0.592 & 0.702 & \textbf{0.587} & \textbf{0.694} & 0.463 & 0.480 & \textbf{0.430} & \textbf{0.432} & \textbf{0.414} & \textbf{0.414} & 0.415 & 0.415 \\
										&  & 336 & 0.447 & 0.456 & \textbf{0.434} & \textbf{0.439} & 0.582 & 0.681 & \textbf{0.579} & \textbf{0.679} & 0.462 & 0.471 & \textbf{0.448} & \textbf{0.457} & 0.440 & 0.445 & \textbf{0.439} & \textbf{0.444} \\
										&  & 720 & 0.483 & 0.503 & \textbf{0.469} & \textbf{0.469} & 0.649 & 0.838 & \textbf{0.640} & \textbf{0.812} & 0.546 & 0.581 & \textbf{0.477} & \textbf{0.475} & 0.471 & 0.473 & \textbf{0.463} & \textbf{0.462} \\ \cmidrule{2-19}
										& \multirow{4}{*}{\rotatebox[origin=c]{90}{\textbf{Dish-ts}}} & 96 & 0.419 & 0.411 & \textbf{0.411} & \textbf{0.398} & 0.755 & 0.962 & \textbf{0.575} & \textbf{0.626} & 0.446 & \textbf{0.433} & \textbf{0.444} & 0.436 & 0.415 & \textbf{0.396} & \textbf{0.413} & 0.402 \\
										&  & 192 & 0.469 & 0.485 & \textbf{0.435} & \textbf{0.437} & \textbf{0.746} & \textbf{0.965} & 0.747 & 0.997 & 0.466 & 0.467 & \textbf{0.454} & \textbf{0.454} & 0.442 & \textbf{0.440} & \textbf{0.439} & 0.443 \\
										&  & 336 & 0.474 & 0.486 & \textbf{0.458} & \textbf{0.470} & 0.878 & 1.157 & \textbf{0.797} & \textbf{1.125} & 0.512 & 0.554 & \textbf{0.508} & \textbf{0.513} & \textbf{0.454} & \textbf{0.465} & 0.466 & 0.484 \\
										&  & 720 & 0.536 & 0.558 & \textbf{0.505} & \textbf{0.517} & 0.811 & 1.164 & \textbf{0.765} & \textbf{1.041} & 0.567 & 0.596 & \textbf{0.561} & \textbf{0.577} & 0.556 & 0.609 & \textbf{0.504} & \textbf{0.516} \\ \cmidrule{2-19}
										& \multirow{4}{*}{\rotatebox[origin=c]{90}{\textbf{SAN}}} & 96 & \textbf{0.414} & \textbf{0.411} & \textbf{0.414} & 0.414 & 0.506 & 0.564 & \textbf{0.481} & \textbf{0.535} & 0.422 & \textbf{0.415} & \textbf{0.419} & 0.420 & 0.498 & 0.566 & \textbf{0.426} & \textbf{0.434} \\
										&  & 192 & 0.435 & 0.451 & \textbf{0.434} & \textbf{0.450} & 0.507 & 0.569 & \textbf{0.491} & \textbf{0.536} & \textbf{0.447} & \textbf{0.457} & \textbf{0.447} & 0.467 & 0.491 & 0.563 & \textbf{0.449} & \textbf{0.472} \\
										&  & 336 & \textbf{0.446} & 0.467 & \textbf{0.446} & \textbf{0.466} & 0.545 & \textbf{0.625} & \textbf{0.539} & 0.640 & \textbf{0.460} & \textbf{0.472} & 0.462 & 0.492 & 0.499 & 0.562 & \textbf{0.458} & \textbf{0.488} \\
										&  & 720 & 0.510 & 0.560 & \textbf{0.493} & \textbf{0.523} & 0.611 & 0.697 & \textbf{0.602} & \textbf{0.686} & 0.535 & 0.564 & \textbf{0.486} & \textbf{0.501} & 0.528 & 0.564 & \textbf{0.525} & \textbf{0.560} \\ \cmidrule{2-19}
										& \multirow{4}{*}{\rotatebox[origin=c]{90}{\textbf{FAN}}} & 96 & 0.422 & 0.407 & \textbf{0.418} & \textbf{0.401} & 0.451 & \textbf{0.432} & \textbf{0.439} & \textbf{0.432} & 0.424 & 0.406 & \textbf{0.420} & \textbf{0.404} & \textbf{0.426} & \textbf{0.408} & 0.433 & 0.414 \\
										&  & 192 & 0.453 & 0.453 & \textbf{0.450} & \textbf{0.451} & 0.494 & 0.499 & \textbf{0.482} & \textbf{0.497} & 0.460 & 0.461 & \textbf{0.452} & \textbf{0.451} & 0.472 & 0.468 & \textbf{0.460} & \textbf{0.464} \\
										&  & 336 & 0.483 & 0.496 & \textbf{0.476} & \textbf{0.487} & 0.525 & 0.550 & \textbf{0.524} & \textbf{0.549} & 0.492 & 0.511 & \textbf{0.479} & \textbf{0.489} & \textbf{0.474} & \textbf{0.489} & 0.478 & 0.500 \\
										&  & 720 & 0.554 & \textbf{0.571} & \textbf{0.549} & 0.574 & 0.607 & 0.652 & \textbf{0.577} & \textbf{0.638} & 0.574 & 0.611 & \textbf{0.556} & \textbf{0.576} & 0.567 & 0.623 & \textbf{0.549} & \textbf{0.573} \\ 
										
										\midrule[1.2pt]
										
										\multirow{20}{*}{\rotatebox[origin=c]{90}{\textbf{ETTh2}}}
										& \multirow{4}{*}{\rotatebox[origin=c]{90}{\textbf{Nones}}} & 96 & 0.381 & 0.331 & \textbf{0.365} & \textbf{0.315} & 1.569 & 3.349 & \textbf{0.869} & \textbf{1.251} & 0.602 & 0.685 & \textbf{0.379} & \textbf{0.327} & \textbf{0.394} & \textbf{0.349} & 0.398 & 0.353 \\
										&  & 192 & 0.462 & 0.446 & \textbf{0.427} & \textbf{0.413} & 1.322 & 2.577 & \textbf{1.168} & \textbf{2.340} & 0.684 & 0.859 & \textbf{0.473} & \textbf{0.466} & \textbf{0.451} & 0.438 & \textbf{0.451} & \textbf{0.437} \\
										&  & 336 & 0.501 & 0.507 & \textbf{0.484} & \textbf{0.477} & 1.324 & 2.595 & \textbf{1.281} & \textbf{2.540} & 0.641 & 0.764 & \textbf{0.519} & \textbf{0.540} & 0.500 & 0.515 & \textbf{0.490} & \textbf{0.497} \\
										&  & 720 & 0.671 & 0.850 & \textbf{0.607} & \textbf{0.739} & 1.518 & 3.013 & \textbf{1.456} & \textbf{2.945} & 0.816 & 1.146 & \textbf{0.602} & \textbf{0.697} & 0.622 & 0.760 & \textbf{0.598} & \textbf{0.705} \\ \cmidrule{2-19}
										& \multirow{4}{*}{\rotatebox[origin=c]{90}{\textbf{RevIN}}} & 96 & 0.368 & 0.316 & \textbf{0.354} & \textbf{0.306} & 0.516 & 0.534 & \textbf{0.476} & \textbf{0.484} & 0.382 & 0.334 & \textbf{0.364} & \textbf{0.311} & \textbf{0.370} & 0.323 & \textbf{0.370} & \textbf{0.321} \\
										&  & 192 & 0.417 & 0.384 & \textbf{0.405} & \textbf{0.383} & 0.556 & 0.650 & \textbf{0.524} & \textbf{0.574} & 0.440 & 0.431 & \textbf{0.413} & \textbf{0.383} & \textbf{0.413} & \textbf{0.385} & \textbf{0.413} & \textbf{0.385} \\
										&  & 336 & 0.443 & \textbf{0.415} & \textbf{0.435} & 0.417 & 0.540 & 0.604 & \textbf{0.487} & \textbf{0.492} & 0.451 & 0.440 & \textbf{0.443} & \textbf{0.424} & \textbf{0.438} & 0.413 & 0.443 & \textbf{0.411} \\
										&  & 720 & \textbf{0.472} & \textbf{0.455} & \textbf{0.472} & \textbf{0.455} & 0.646 & 0.845 & \textbf{0.530} & \textbf{0.575} & 0.484 & 0.480 & \textbf{0.470} & \textbf{0.456} & \textbf{0.486} & \textbf{0.477} & \textbf{0.486} & 0.764 \\ \cmidrule{2-19}
										& \multirow{4}{*}{\rotatebox[origin=c]{90}{\textbf{Dish-ts}}} & 96 & \textbf{0.368} & 0.316 & 0.371 & \textbf{0.312} & 1.389 & 3.795 & \textbf{0.666} & \textbf{0.954} & 0.414 & 0.365 & \textbf{0.408} & \textbf{0.364} & 0.470 & 0.472 & \textbf{0.394} & \textbf{0.353} \\
										&  & 192 & 0.449 & 0.414 & \textbf{0.418} & \textbf{0.378} & 1.183 & 2.626 & \textbf{0.803} & \textbf{1.367} & 0.508 & 0.519 & \textbf{0.449} & \textbf{0.429} & 0.497 & \textbf{0.512} & \textbf{0.487} & 0.519 \\
										&  & 336 & 0.486 & 0.465 & \textbf{0.458} & \textbf{0.436} & 1.348 & 3.385 & \textbf{0.943} & \textbf{1.834} & 0.531 & 0.551 & \textbf{0.499} & \textbf{0.507} & 0.677 & 0.896 & \textbf{0.580} & \textbf{0.686} \\
										&  & 720 & 0.599 & 0.667 & \textbf{0.530} & \textbf{0.549} & 1.366 & 3.238 & \textbf{0.770} & \textbf{1.117} & 0.696 & 0.914 & \textbf{0.584} & \textbf{0.687} & 0.878 & 1.851 & \textbf{0.673} & \textbf{0.889} \\ \cmidrule{2-19}
										& \multirow{4}{*}{\rotatebox[origin=c]{90}{\textbf{SAN}}} & 96 & 0.376 & 0.325 & \textbf{0.369} & \textbf{0.316} & 0.516 & 0.541 & \textbf{0.390} & \textbf{0.344} & 0.374 & \textbf{0.325} & \textbf{0.288} & 0.328 & 0.462 & 0.424 & \textbf{0.401} & \textbf{0.348} \\
										&  & 192 & 0.425 & 0.391 & \textbf{0.422} & \textbf{0.390} & 0.633 & 0.734 & \textbf{0.440} & \textbf{0.428} & \textbf{0.424} & \textbf{0.401} & 0.428 & 0.403 & 0.534 & 0.570 & \textbf{0.459} & \textbf{0.444} \\
										&  & 336 & 0.455 & \textbf{0.426} & \textbf{0.454} & 0.435 & 0.698 & 0.836 & \textbf{0.462} & \textbf{0.455} & \textbf{0.447} & \textbf{0.427} & 0.453 & 0.434 & 0.564 & 0.618 & \textbf{0.521} & \textbf{0.540} \\
										&  & 720 & 0.497 & 0.480 & \textbf{0.488} & \textbf{0.477} & 1.024 & 1.646 & \textbf{0.497} & \textbf{0.523} & 0.507 & 0.520 & \textbf{0.488} & \textbf{0.476} & 0.676 & 0.863 & \textbf{0.560} & \textbf{0.579} \\\cmidrule{2-19}
										& \multirow{4}{*}{\rotatebox[origin=c]{90}{\textbf{FAN}}} & 96 & 0.380 & 0.328 & \textbf{0.379} & \textbf{0.326} & 0.446 & 0.400 & \textbf{0.411} & \textbf{0.370} & 0.387 & 0.334 & \textbf{0.384} & \textbf{0.333} & 0.400 & 0.354 & \textbf{0.392} & \textbf{0.341} \\
										&  & 192 & \textbf{0.431} & \textbf{0.402} & 0.434 & 0.408 & 0.521 & 0.533 & \textbf{0.460} & \textbf{0.445} & 0.442 & 0.419 & \textbf{0.439} & \textbf{0.418} & 0.445 & 0.420 & \textbf{0.434} & \textbf{0.409} \\
										&  & 336 & 0.483 & 0.486 & \textbf{0.469} & \textbf{0.460} & 0.540 & 0.547 & \textbf{0.470} & \textbf{0.448} & 0.488 & 0.487 & \textbf{0.483} & \textbf{0.478} & 0.471 & \textbf{0.456} & \textbf{0.470} & 0.457 \\
										&  & 720 & 0.615 & 0.732 & \textbf{0.597} & \textbf{0.686} & 0.751 & 1.014 & \textbf{0.616} & \textbf{0.734} & 0.649 & 0.806 & \textbf{0.626} & \textbf{0.759} & \textbf{0.585} & \textbf{0.638} & 0.594 & 0.685 \\
										
										\midrule
										\bottomrule[1.2pt]
									\end{tabular}
								}
								\caption{Part I of the detailed results of the main experiment}
								\label{main1}
							\end{table*}

							\begin{table*}[t]
								\centering 		 		 		
								
								\small
								\renewcommand\arraystretch{0.75}
									\setlength{\tabcolsep}{4pt}{
										\begin{tabular}{c|c|c|cc|cc|cc|cc|cc|cc|cc|cc}
											\toprule[1.2pt]
											\multicolumn{3}{c|}{\textbf{Methods}}  & \multicolumn{4}{c|}{\textbf{CATS}} & \multicolumn{4}{c|}{\textbf{Informer}} & \multicolumn{4}{c|}{\textbf{iTransformer}} & \multicolumn{4}{c}{\textbf{SparseTSF}}\\ 
											\midrule
											\multicolumn{3}{c|}{\textbf{Affine}} & && \multicolumn{2}{c|}{\textbf{+APT}}&&& \multicolumn{2}{c|}{\textbf{+APT}}&&& \multicolumn{2}{c|}{\textbf{+APT}}&&& \multicolumn{2}{c}{\textbf{+APT}}\\
											\midrule
											\multicolumn{3}{c|}{\textbf{Metrics}} & MAE & MSE & MAE & MSE& MAE & MSE& MAE & MSE& MAE & MSE& MAE & MSE& MAE & MSE& MAE & MSE\\
											\midrule[1.2pt]
											\multirow{20}{*}{\rotatebox[origin=c]{90}{\textbf{Exchange}}}
											& \multirow{4}{*}{\rotatebox[origin=c]{90}{\textbf{Nones}}} & 96 & 0.335 & 0.233 & \textbf{0.246} & \textbf{0.105} & 0.896 & 1.255 & \textbf{0.879} & \textbf{1.215} & 0.443 & 0.374 & \textbf{0.283} & \textbf{0.152} & 0.265 & 0.124 & \textbf{0.256} & \textbf{0.114} \\
											&  & 192 & 0.435 & 0.371 & \textbf{0.319} & \textbf{0.181} & 1.016 & 1.719 & \textbf{0.961} & \textbf{1.364} & 0.535 & 0.493 & \textbf{0.462} & \textbf{0.351} & 0.327 & \textbf{0.184} & \textbf{0.324} & 0.186 \\
											&  & 336 & 0.514 & 0.466 & \textbf{0.410} & \textbf{0.285} & 1.058 & 1.855 & \textbf{0.976} & \textbf{1.510} & 0.661 & 0.704 & \textbf{0.502} & \textbf{0.444} & 0.448 & 0.341 & \textbf{0.432} & \textbf{0.316} \\
											&  & 720 & 1.184 & 2.483 & \textbf{0.686} & \textbf{0.695} & 1.015 & 1.732 & \textbf{0.863} & \textbf{1.258} & 0.981 & 1.560 & \textbf{0.956} & \textbf{1.450} & 0.664 & 0.757 & \textbf{0.662} & \textbf{0.754} \\ \cmidrule{2-19}
											& \multirow{4}{*}{\rotatebox[origin=c]{90}{\textbf{RevIN}}} & 96 & 0.213 & 0.090 & \textbf{0.199} & \textbf{0.079} & 0.456 & 0.354 & \textbf{0.344} & \textbf{0.196} & 0.235 & 0.103 & \textbf{0.213} & \textbf{0.089} & 0.235 & 0.105 & \textbf{0.230} & \textbf{0.101} \\
											&  & 192 & \textbf{0.299} & \textbf{0.166} & \textbf{0.299} & 0.170 & 0.500 & 0.402 & \textbf{0.480} & \textbf{0.386} & 0.350 & 0.221 & \textbf{0.328} & \textbf{0.197} & 0.344 & 0.221 & \textbf{0.330} & \textbf{0.206} \\
											&  & 336 & 0.437 & 0.352 & \textbf{0.423} & \textbf{0.331} & \textbf{0.677} & \textbf{0.708} & 0.686 & 0.711 & 0.493 & 0.423 & \textbf{0.459} & \textbf{0.378} & 0.499 & 0.434 & \textbf{0.460} & \textbf{0.393} \\
											&  & 720 & 0.789 & 0.997 & \textbf{0.774} & \textbf{0.943} & 0.734 & 0.885 & \textbf{0.565} & \textbf{0.546} & 0.852 & 1.124 & \textbf{0.787} & \textbf{1.001} & 0.844 & 1.155 & \textbf{0.732} & \textbf{0.846} \\\cmidrule{2-19}
											& \multirow{4}{*}{\rotatebox[origin=c]{90}{\textbf{Dish-ts}}} & 96 & 0.270 & 0.114 & \textbf{0.256} & \textbf{0.105} & 0.425 & 0.308 & \textbf{0.423} & \textbf{0.307} & 0.293 & 0.155 & \textbf{0.263} & \textbf{0.111} & 0.301 & 0.155 & \textbf{0.284} & \textbf{0.128} \\
											&  & 192 & 0.512 & 0.428 & \textbf{0.353} & \textbf{0.211} & \textbf{0.646} & 0.735 & 0.648 & \textbf{0.726} & 0.386 & 0.265 & \textbf{0.362} & \textbf{0.234} & \textbf{0.417} & 0.325 & 0.420 & \textbf{0.298} \\
											&  & 336 & \textbf{0.491} & \textbf{0.388} & 0.517 & 0.406 & 0.940 & 1.428 & \textbf{0.736} & \textbf{0.831} & 0.552 & 0.508 & \textbf{0.536} & \textbf{0.457} & 0.642 & 0.632 & \textbf{0.537} & \textbf{0.469} \\
											&  & 720 & 0.974 & 2.165 & \textbf{0.847} & \textbf{1.150} & 1.709 & 5.947 & \textbf{1.008} & \textbf{2.150} & 0.806 & 1.054 & \textbf{0.745} & \textbf{0.776} & 0.800 & 1.012 & \textbf{0.768} & \textbf{1.009} \\ \cmidrule{2-19}
											& \multirow{4}{*}{\rotatebox[origin=c]{90}{\textbf{SAN}}} & 96 & 0.242 & 0.112 & \textbf{0.202} & \textbf{0.078} & 0.254 & 0.119 & \textbf{0.212} & \textbf{0.081} & 0.245 & 0.118 & \textbf{0.204} & \textbf{0.080} & 0.219 & 0.087 & \textbf{0.212} & \textbf{0.083} \\
											&  & 192 & 0.338 & 0.212 & \textbf{0.296} & \textbf{0.163} & 0.396 & 0.300 & \textbf{0.298} & \textbf{0.167} & 0.396 & 0.307 & \textbf{0.298} & \textbf{0.167} & 0.312 & \textbf{0.169} & \textbf{0.302} & 0.171 \\
											&  & 336 & 0.500 & 0.474 & \textbf{0.414} & \textbf{0.314} & 0.484 & 0.394 & \textbf{0.420} & \textbf{0.321} & 0.515 & 0.473 & \textbf{0.471} & \textbf{0.396} & \textbf{0.419} & \textbf{0.305} & 0.422 & 0.324 \\
											&  & 720 & 0.853 & 1.287 & \textbf{0.748} & \textbf{0.935} & 0.696 & 0.827 & \textbf{0.685} & \textbf{0.782} & \textbf{0.827} & \textbf{1.098} & 0.857 & 1.149 & 0.720 & \textbf{0.839} & \textbf{0.694} & 0.844 \\ \cmidrule{2-19}
											& \multirow{4}{*}{\rotatebox[origin=c]{90}{\textbf{FAN}}} & 96 & 0.301 & 0.152 & \textbf{0.261} & \textbf{0.130} & 0.330 & 0.194 & \textbf{0.308} & \textbf{0.183} & 0.278 & 0.133 & \textbf{0.241} & \textbf{0.104} & \textbf{0.247} & \textbf{0.109} & \textbf{0.247} & \textbf{0.109} \\
											&  & 192 & 0.381 & 0.248 & \textbf{0.371} & \textbf{0.238} & 0.487 & 0.420 & \textbf{0.462} & \textbf{0.372} & 0.412 & 0.285 & \textbf{0.394} & \textbf{0.264} & 0.372 & 0.251 & \textbf{0.349} & \textbf{0.213} \\
											&  & 336 & \textbf{0.518} & \textbf{0.452} & 0.531 & 0.463 & 0.631 & \textbf{0.683} & \textbf{0.626} & 0.686 & 0.594 & 0.574 & \textbf{0.548} & \textbf{0.494} & 0.564 & 0.553 & \textbf{0.540} & \textbf{0.493} \\
											&  & 720 & \textbf{0.764} & \textbf{0.980} & 0.765 & 0.997 & 0.801 & 1.068 & \textbf{0.738} & \textbf{0.873} & 0.768 & 1.018 & \textbf{0.759} & \textbf{0.931} & \textbf{0.710} & \textbf{0.859} & 0.713 & 0.861 \\
											
											\midrule[1.2pt]
											
											\multirow{20}{*}{\rotatebox[origin=c]{90}{\textbf{Traffic}}}
											& \multirow{4}{*}{\rotatebox[origin=c]{90}{\textbf{Nones}}} & 96 & \textbf{0.269} & 0.509 & 0.271 & \textbf{0.503} & 0.432 & 0.832 & \textbf{0.397} & \textbf{0.764} & 0.550 & 0.876 & \textbf{0.429} & \textbf{0.624} & \textbf{0.287} & \textbf{0.423} & \textbf{0.287} & \textbf{0.423} \\
											&  & 192 & 0.284 & 0.549 & \textbf{0.281} & \textbf{0.510} & 0.392 & 0.772 & \textbf{0.386} & \textbf{0.759} & 0.560 & 0.909 & \textbf{0.416} & \textbf{0.617} & \textbf{0.292} & \textbf{0.435} & \textbf{0.292} & \textbf{0.435} \\
											&  & 336 & \textbf{0.290} & 0.563 & 0.292 & \textbf{0.542} & 0.401 & 0.787 & \textbf{0.397} & \textbf{0.780} & 0.597 & \textbf{1.002} & \textbf{0.432} & 1.178 & 0.298 & \textbf{0.447} & \textbf{0.296} & \textbf{0.447} \\
											&  & 720 & 0.310 & 0.594 & \textbf{0.302} & \textbf{0.565} & 0.474 & 0.928 & \textbf{0.427} & \textbf{0.816} & 0.668 & 1.156 & \textbf{0.436} & \textbf{0.903} & 0.316 & 0.469 & \textbf{0.311} & \textbf{0.468} \\ \cmidrule{2-19}
											& \multirow{4}{*}{\rotatebox[origin=c]{90}{\textbf{RevIN}}} & 96 & \textbf{0.263} & \textbf{0.382} & 0.264 & 0.386 & 0.395 & 0.735 & \textbf{0.360} & \textbf{0.689} & \textbf{0.270} & \textbf{0.375} & 0.276 & 0.380 & \textbf{0.285} & \textbf{0.424} & \textbf{0.285} & \textbf{0.424} \\
											&  & 192 & 0.283 & 0.418 & \textbf{0.274} & \textbf{0.407} & 0.489 & 0.941 & \textbf{0.374} & \textbf{0.713} & \textbf{0.281} & 0.403 & 0.287 & \textbf{0.401} & \textbf{0.290} & 0.437 & \textbf{0.290} & \textbf{0.436} \\
											&  & 336 & 0.287 & 0.431 & \textbf{0.277} & \textbf{0.414} & 0.430 & 0.820 & \textbf{0.427} & \textbf{0.814} & \textbf{0.291} & \textbf{0.420} & 0.294 & 0.422 & 0.296 & \textbf{0.448} & \textbf{0.295} & 0.449 \\
											&  & 720 & \textbf{0.305} & \textbf{0.454} & 0.306 & 0.458 & 0.513 & 1.000 & \textbf{0.416} & \textbf{0.794} & \textbf{0.311} & \textbf{0.449} & 0.315 & 0.456 & 0.311 & \textbf{0.470} & \textbf{0.310} & \textbf{0.470} \\ \cmidrule{2-19}
											& \multirow{4}{*}{\rotatebox[origin=c]{90}{\textbf{Dish-ts}}} & 96 & 0.279 & 0.411 & \textbf{0.263} & \textbf{0.387} & 0.385 & 0.710 & \textbf{0.357} & \textbf{0.683} & \textbf{0.288} & \textbf{0.402} & 0.293 & 0.403 & 0.302 & 0.440 & \textbf{0.299} & \textbf{0.436} \\
											&  & 192 & 0.281 & 0.421 & \textbf{0.272} & \textbf{0.408} & 0.388 & 0.732 & \textbf{0.373} & \textbf{0.702} & \textbf{0.295} & \textbf{0.413} & 0.299 & 0.414 & 0.308 & 0.454 & \textbf{0.304} & \textbf{0.448} \\
											&  & 336 & 0.300 & 0.450 & \textbf{0.280} & \textbf{0.424} & 0.472 & 0.925 & \textbf{0.436} & \textbf{0.835} & \textbf{0.308} & \textbf{0.434} & 0.313 & 0.440 & 0.314 & 0.462 & \textbf{0.310} & \textbf{0.459} \\
											&  & 720 & 0.312 & 0.472 & \textbf{0.297} & \textbf{0.450} & 0.536 & 1.077 & \textbf{0.447} & \textbf{0.825} & \textbf{0.330} & 0.467 & 0.332 & \textbf{0.465} & 0.331 & 0.488 & \textbf{0.329} & \textbf{0.484} \\ \cmidrule{2-19}
											& \multirow{4}{*}{\rotatebox[origin=c]{90}{\textbf{SAN}}} & 96 & \textbf{0.275} & \textbf{0.395} & 0.281 & 0.405 & 0.355 & 0.623 & \textbf{0.339} & \textbf{0.608} & \textbf{0.278} & \textbf{0.389} & 0.281 & 0.390 & 0.420 & 0.661 & \textbf{0.292} & \textbf{0.419} \\
											&  & 192 & \textbf{0.286} & \textbf{0.421} & 0.290 & 0.427 & 0.403 & 0.715 & \textbf{0.377} & \textbf{0.673} & \textbf{0.288} & \textbf{0.416} & 0.291 & 0.418 & 0.361 & 0.557 & \textbf{0.346} & \textbf{0.535} \\
											&  & 336 & 0.294 & 0.440 & \textbf{0.287} & \textbf{0.439} & 0.397 & 0.727 & \textbf{0.373} & \textbf{0.683} & \textbf{0.296} & \textbf{0.436} & 0.300 & 0.440 & 0.414 & 0.670 & \textbf{0.409} & \textbf{0.659} \\
											&  & 720 & 0.312 & \textbf{0.472} & \textbf{0.306} & 0.476 & 0.521 & 0.947 & \textbf{0.485} & \textbf{0.899} & \textbf{0.313} & \textbf{0.469} & 0.315 & 0.477 & 0.433 & 0.716 & \textbf{0.404} & \textbf{0.662} \\ \cmidrule{2-19}
											& \multirow{4}{*}{\rotatebox[origin=c]{90}{\textbf{FAN}}} & 96 & 0.290 & 0.414 & \textbf{0.288} & \textbf{0.412} & 0.290 & 0.418 & \textbf{0.281} & \textbf{0.413} & 0.313 & 0.429 & \textbf{0.304} & \textbf{0.422} & 0.283 & 0.395 & \textbf{0.277} & \textbf{0.393} \\
											&  & 192 & 0.306 & 0.439 & \textbf{0.291} & \textbf{0.421} & 0.304 & 0.440 & \textbf{0.292} & \textbf{0.432} & 0.334 & 0.453 & \textbf{0.310} & \textbf{0.435} & 0.293 & 0.419 & \textbf{0.291} & \textbf{0.417} \\
											&  & 336 & 0.315 & 0.455 & \textbf{0.297} & \textbf{0.435} & 0.317 & 0.459 & \textbf{0.302} & \textbf{0.448} & 0.340 & 0.468 & \textbf{0.324} & \textbf{0.455} & \textbf{0.300} & \textbf{0.433} & 0.302 & 0.437 \\
											&  & 720 & 0.352 & 0.507 & \textbf{0.330} & \textbf{0.484} & 0.350 & 0.510 & \textbf{0.329} & \textbf{0.486} & 0.375 & 0.520 & \textbf{0.360} & \textbf{0.504} & 0.330 & 0.481 & \textbf{0.324} & \textbf{0.478} \\
											
											\midrule[1.2pt]
											
											\multirow{20}{*}{\rotatebox[origin=c]{90}{\textbf{Weather}}}
											& \multirow{4}{*}{\rotatebox[origin=c]{90}{\textbf{Nones}}} & 96 & 0.203 & 0.147 & \textbf{0.195} & \textbf{0.145} & 0.243 & 0.184 & \textbf{0.218} & \textbf{0.164} & 0.224 & \textbf{0.163} & \textbf{0.220} & \textbf{0.163} & 0.258 & 0.192 & \textbf{0.253} & \textbf{0.190} \\
											&  & 192 & 0.251 & 0.195 & \textbf{0.244} & \textbf{0.188} & 0.342 & 0.327 & \textbf{0.312} & \textbf{0.255} & 0.310 & 0.267 & \textbf{0.266} & \textbf{0.209} & 0.282 & 0.227 & \textbf{0.280} & \textbf{0.226} \\
											&  & 336 & 0.294 & \textbf{0.244} & \textbf{0.287} & 0.245 & 0.519 & 0.576 & \textbf{0.315} & \textbf{0.270} & 0.314 & 0.261 & \textbf{0.293} & \textbf{0.252} & 0.320 & 0.270 & \textbf{0.313} & \textbf{0.266} \\
											&  & 720 & 0.374 & 0.329 & \textbf{0.348} & \textbf{0.315} & 0.482 & 0.516 & \textbf{0.401} & \textbf{0.381} & 0.361 & 0.333 & \textbf{0.360} & \textbf{0.330} & 0.367 & \textbf{0.329} & \textbf{0.348} & 0.359 \\ \cmidrule{2-19}
											& \multirow{4}{*}{\rotatebox[origin=c]{90}{\textbf{RevIN}}} & 96 & 0.193 & 0.145 & \textbf{0.192} & \textbf{0.144} & 0.214 & 0.175 & \textbf{0.213} & \textbf{0.167} & \textbf{0.206} & \textbf{0.158} & 0.208 & \textbf{0.158} & \textbf{0.235} & \textbf{0.185} & \textbf{0.235} & \textbf{0.185} \\
											&  & 192 & \textbf{0.233} & \textbf{0.184} & 0.234 & 0.189 & 0.283 & 0.258 & \textbf{0.257} & \textbf{0.214} & \textbf{0.245} & \textbf{0.200} & 0.248 & 0.201 & 0.265 & 0.224 & \textbf{0.263} & \textbf{0.223} \\
											&  & 336 & 0.274 & \textbf{0.236} & \textbf{0.273} & 0.238 & 0.311 & 0.307 & \textbf{0.290} & \textbf{0.263} & \textbf{0.284} & 0.251 & \textbf{0.284} & \textbf{0.250} & 0.294 & \textbf{0.267} & \textbf{0.293} & \textbf{0.267} \\
											&  & 720 & \textbf{0.331} & \textbf{0.316} & 0.336 & 0.325 & 0.366 & 0.375 & \textbf{0.335} & \textbf{0.329} & \textbf{0.333} & \textbf{0.323} & 0.335 & \textbf{0.323} & 0.339 & \textbf{0.333} & \textbf{0.338} & \textbf{0.333} \\ \cmidrule{2-19}
											& \multirow{4}{*}{\rotatebox[origin=c]{90}{\textbf{Dish-ts}}} & 96 & 0.216 & 0.151 & \textbf{0.212} & \textbf{0.150} & 0.231 & 0.188 & \textbf{0.228} & \textbf{0.187} & 0.238 & 0.190 & \textbf{0.220} & \textbf{0.162} & \textbf{0.234} & \textbf{0.168} & 0.239 & 0.169 \\
											&  & 192 & 0.255 & \textbf{0.191} & \textbf{0.250} & \textbf{0.191} & 0.310 & 0.277 & \textbf{0.285} & \textbf{0.266} & \textbf{0.255} & \textbf{0.199} & 0.261 & 0.206 & 0.281 & \textbf{0.210} & \textbf{0.274} & \textbf{0.210} \\
											&  & 336 & \textbf{0.290} & \textbf{0.240} & 0.291 & 0.242 & 0.334 & 0.323 & \textbf{0.300} & \textbf{0.272} & 0.309 & 0.264 & \textbf{0.300} & \textbf{0.256} & 0.320 & \textbf{0.257} & \textbf{0.315} & \textbf{0.257} \\
											&  & 720 & \textbf{0.349} & \textbf{0.317} & 0.352 & 0.321 & 0.388 & 0.394 & \textbf{0.369} & \textbf{0.381} & 0.382 & 0.378 & \textbf{0.358} & \textbf{0.336} & 0.370 & \textbf{0.325} & \textbf{0.366} & \textbf{0.325} \\ \cmidrule{2-19}
											& \multirow{4}{*}{\rotatebox[origin=c]{90}{\textbf{SAN}}} & 96 & \textbf{0.209} & \textbf{0.148} & 0.215 & 0.151 & 0.228 & 0.181 & \textbf{0.210} & \textbf{0.161} & 0.213 & 0.154 & \textbf{0.211} & \textbf{0.153} & \textbf{0.213} & \textbf{0.150} & \textbf{0.213} & 0.151 \\
											&  & 192 & \textbf{0.258} & \textbf{0.195} & \textbf{0.258} & 0.196 & \textbf{0.253} & \textbf{0.214} & 0.257 & 0.217 & \textbf{0.253} & 0.198 & 0.254 & \textbf{0.195} & 0.258 & 0.195 & \textbf{0.256} & \textbf{0.194} \\
											&  & 336 & 0.294 & \textbf{0.242} & \textbf{0.293} & \textbf{0.242} & 0.304 & 0.282 & \textbf{0.299} & \textbf{0.265} & \textbf{0.281} & \textbf{0.245} & 0.291 & 0.247 & \textbf{0.294} & \textbf{0.243} & 0.297 & 0.247 \\
											&  & 720 & \textbf{0.347} & \textbf{0.316} & 0.352 & 0.319 & 0.359 & 0.366 & \textbf{0.339} & \textbf{0.336} & \textbf{0.336} & \textbf{0.314} & 0.350 & 0.347 & 0.349 & \textbf{0.316} & \textbf{0.343} & 0.317 \\ \cmidrule{2-19}
											& \multirow{4}{*}{\rotatebox[origin=c]{90}{\textbf{FAN}}} & 96 & \textbf{0.208} & \textbf{0.153} & 0.213 & 0.157 & 0.229 & 0.172 & \textbf{0.207} & \textbf{0.157} & 0.218 & 0.158 & \textbf{0.209} & \textbf{0.154} & \textbf{0.208} & \textbf{0.152} & 0.214 & 0.154 \\
											&  & 192 & 0.261 & \textbf{0.199} & \textbf{0.260} & 0.205 & \textbf{0.257} & \textbf{0.207} & 0.258 & 0.208 & 0.264 & 0.203 & \textbf{0.253} & \textbf{0.198} & 0.252 & \textbf{0.197} & \textbf{0.251} & 0.198 \\
											&  & 336 & \textbf{0.298} & \textbf{0.249} & 0.303 & 0.252 & 0.295 & 0.257 & \textbf{0.293} & \textbf{0.249} & 0.307 & 0.258 & \textbf{0.294} & \textbf{0.248} & \textbf{0.296} & \textbf{0.247} & \textbf{0.296} & \textbf{0.247} \\
											&  & 720 & \textbf{0.356} & 0.325 & 0.357 & \textbf{0.321} & \textbf{0.359} & \textbf{0.348} & 0.364 & 0.364 & 0.362 & 0.326 & \textbf{0.347} & \textbf{0.318} & \textbf{0.346} & \textbf{0.317} & 0.353 & 0.320 \\
											
											\midrule
											\bottomrule[1.2pt]
										\end{tabular}
									}
									\caption{Part II of the detailed results of the main experiment}
									\label{main2}
								\end{table*}
								
								\begin{figure*}[t]
									\centering
									\includegraphics[width=0.8\textwidth]{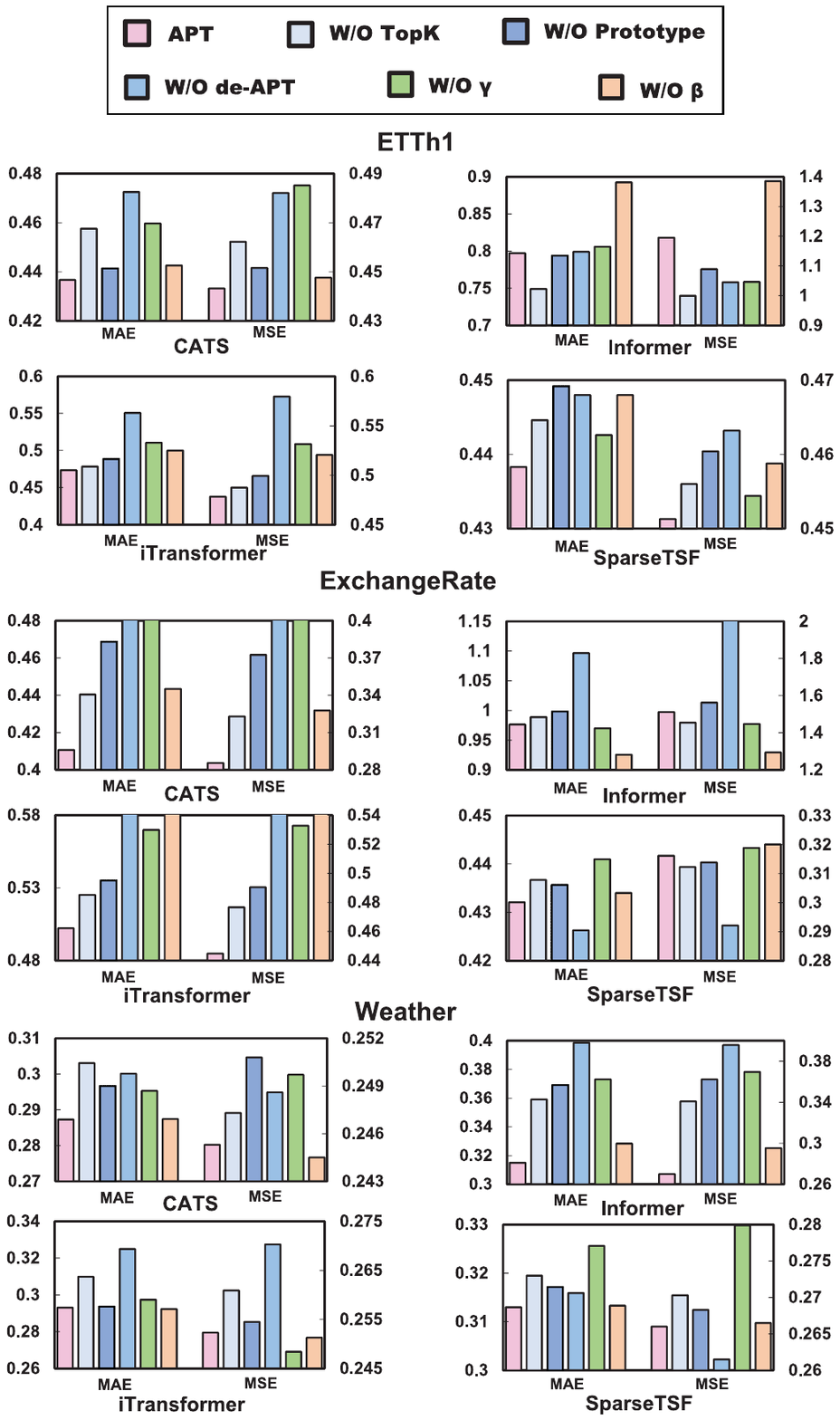}
									\caption{The ablation study results of APT components in ETTh1, ExchangeRate and Weather, $L=336, H=336$}
									\label{xrall}
								\end{figure*}

								\subsection{Hyper-parameter Sensitivity Study}\label{hyperstudy}
								
								In this section, we discuss the impact of hyperparameters on APT. Given that this section does not have a significant impact compared to ablation studies, we have placed it entirely in the appendix for the comprehensiveness of experiment.
								
								\noindent \textbf{Potential extra Parameter:}
								
								\begin{itemize}
									\item \textbf{+ Double prototype}: Double the number of prototypes assigned to each dataset in Table~\ref{datatable}.
									\item \textbf{+ Identity embedding}:  Add identity embedding $\textbf{\textsl{ID}}\in \mathbb{R}^{C \times D}$ to $\widetilde{\textbf{\textsl{T}}}_{t}$ in Equation~\ref{idemb}, which is considered optional in the main text, to explore whether different affine parameters can be assigned to different channels.
									\item \textbf{+ “Day in Month”}: Add “Day in Month” to the timestamp representation in Equation~\ref{id}. This label is considered an insignificant timestamp used to test the acceptance level of redundancy, as time series rarely generate associations on this timestamp.
								\end{itemize}
								The results of hyper-parameter sensitivity study are shown in Table~\ref{hyper}. Prototype learning is originally introduced to address the few-shot challenge at each timestamp. Increasing the number of prototypes provides more fine-grained temporal representations but also raises the risk of overfitting. This hyper-parameter is dataset-dependent and  a larger prototype library tends to improve performance on ETTh1.
								
								Identity embedding aims to differentiate affine parameters across channels. However, we omit this component to maintain APT's minimal complexity and avoid redundancy with channel-specific mechanisms already present in some backbones. Empirically, it only yields noticeable gains on Exchange and ETTh1 due to their limited channel counts, and affine parameters may struggle to capture complex spatial correlations.
								
								Among timestamp embeddings, “Day in Month” proves less informative than “Day in Week” or “Time in Hour”. The latter two often reflect external factors influenced by diurnal cycles and human activity, whereas “Day in Month” introduces noise and redundancy, degrading APT’s global distributional awareness and consistently reducing performance across all backbones.
								
								\noindent \textbf{Embedding Size} is the most important hyper-parameter for APT intuitively, as it determines the representation space of timestamps and prototypes. We show the impact of different embedding sizes on model performance in Table~\ref{embeddingsize}.
								
								This parameter has minimal impact on APT's performance in most settings, with sensitivity primarily observed on Informer and the Exchange dataset. Larger embedding sizes may lead to overfitting and degrade forecasting accuracy.
								In the main experiments, we recommend a fixed embedding size of 20 across all backbone–normalization combinations for simplicity and consistency.
								
								\subsection{Parameter Count Comparison}\label{paracount}
								We report the parameter counts of each backbone and normalization on ETTh1 in Table~\ref{para}. Among the backbones, SparseTSF is a well-known lightweight linear model. Its patch mechanism results in even fewer parameters than DLinear, with a total below 1K. It explains why APT brings limited improvements when paired with SparseTSF—its minimal capacity may be insufficient to fully utilize the dynamic affine transformations provided by APT.
								
								In contrast, CATS, Informer, and iTransformer each exceed 1M parameters, which is over three orders of magnitude larger than APT. Notably, Informer's parameter count is independent of input length due to channel dependency.
								
								For normalization strategies, RevIN introduces affine parameters twice the number of input channels, yet as shown in Table~\ref{affine1}, these have limited effect on forecasting performance. Dish-TS, despite underperforming, requires more than twice the parameters of APT, while advanced methods like SAN and FAN are 10–100× larger.
								
								Parameter count directly reflects that APT is a lightweight plugin. In addition to the performance improvements brought about by global distribution awareness, it is compatible with any backbone and normalization strategy, and often does not impose any computational burden.
								\subsection{Additional Ablation Results}~\label{addabl}
								\subsubsection{Component Ablation:}We evaluate APT by removing key components. For ease of understanding, we have copied the introduction from the main text below:
								
								\begin{itemize}
									\item\textbf{W/O Top-$k$}: Remove Top-$k$ from Equation~\ref{softmax};
									\item \textbf{W/O Prototype}: Use raw timestamp embeddings without prototype matching;
									\item \textbf{W/O de-APT}: Remove inverse transformation of APT;
									\item \textbf{W/O $\gamma$ or $\beta$}: Remove scaling or bias in affine transformation;;
								\end{itemize}
								
								Compared with the main text, more experimental results are shown in Figure~\ref{xrall}. The main conclusions align with those presented in the main text. Removing the inverse APT transformation imposes an additional burden on the prediction head, which must learn to reconstruct the original value space from APT’s transformed representation. In contrast, de-APT preserves structural symmetry across the pipeline and alleviates this issue.
								
								Top-$k$ weighting, prototype learning, and affine parameters are all mechanisms related to managing distributional risk. Due to the limited priors and lack of manual intervention in deep learning, models are susceptible to risks such as feature oversmoothing, feature collapse, or diminished reliance on affine parameters when supervising learning. These factors are highly dataset- and model-dependent, often introducing randomness into the optimization process.
								
								Top-$k$ and prototype learning are conceptually opposite in representation: Top-$k$ enforces discriminative assignments across timestamps, while prototype learning promotes generalization for underrepresented or highly variable timestamp features. In APT, these two components are balanced to mitigate prediction instability.
								
								In some dataset–backbone combinations, removing the affine bias term $\beta$ occasionally leads to performance gains. Nevertheless, we retain both affine parameters in the main experiments, as such fluctuations are irregular and require extensive manual tuning to validate, which does not align with APT's pursuit of lightweight and flexibility.

								\subsection{Cross-Setting Fine-Tuning}~\label{crosssetting}
								Since APT's parameters are independent of forecasting length, and its distribution shift mitigation is inherently dataset-driven, we further investigate the flexibility and generalization of APT under diverse settings.
								This experiment evaluates APT’s performance under \textbf{cross-backbone} $\boldsymbol{\mathcal{M}}$ , \textbf{cross-normalization} $\boldsymbol{\mathcal{N}}$, and \textbf{cross-length $H$} conditions.
								
								We conduct the study on the ETTh1 dataset by transferring pretrained APT parameters and applying one epoch of fine-tuning for each new setting.
								This setup follows observations from additional experiments, which indicate that zero-shot performance is limited without fine-tuning.
								
								\subsubsection{Cross-Backbone $\boldsymbol{\mathcal{M}}$:} We incorporate APT parameters learned from a source backbone into the target backbone’s forecasting pipeline and perform one epoch of fine-tuning. As shown in Table~\ref{cross}, this consistently improves performance across all settings, often surpassing the results of joint training with APT on the target backbone. These results demonstrate that APT possesses cross-backbone transferability and can directly enhance arbitrary forecasting backbone with pretrained parameters and  minimal fine-tuning.
								
								\subsubsection{Cross-Normalization $\boldsymbol{\mathcal{N}}$:} For iTransformer, we further test transferring APT parameters across normalization strategies. By injecting APT parameters learned under a source normalization into the target normalization pipeline and fine-tuning for one epoch, we again observe results that are comparable to the original configuration. While only a single epoch is used, performance can steadily improve with additional training, mirroring the behavior observed in the cross-backbone setting.
								
								\subsubsection{Cross-Length $H$:} We also attempt to transfer APT across different output lengths, but find that this setup generally fails to reach optimal performance. Fine-tuning in these cases cannot recover the performance of jointly trained APT. We attribute this to distribution shift introduced by varying output lengths, which result in misaligned feature spaces that undermine the generalizability of a length-specific APT model.
								
								In summary, when the output length is consistent, APT demonstrates strong generalization across backbones and normalization methods. This suggests that APT effectively captures global distributional features intrinsic to the dataset, independent of model architecture or normalization strategy. A single epoch of fine-tuning is sufficient to align module interactions and recover performance, highlighting the flexibility and generalization of APT across diverse forecasting configurations.

								\begin{figure*}[t]
									\centering
									\includegraphics[width=0.92\textwidth]{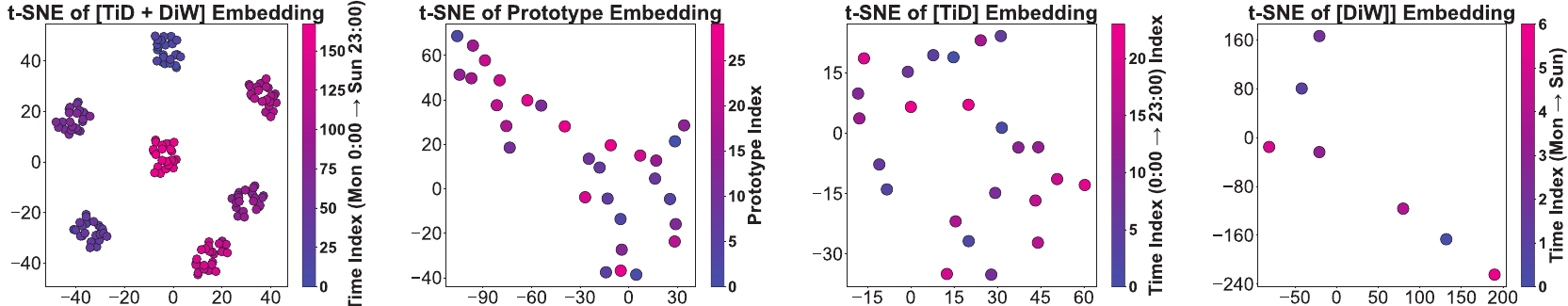}
									\caption{Visualization of APT's embeddings on ECL dataset and iTransformer}
									\label{tsneecl}
								\end{figure*}
								\begin{figure*}[t]
									\centering
									\includegraphics[width=0.92\textwidth]{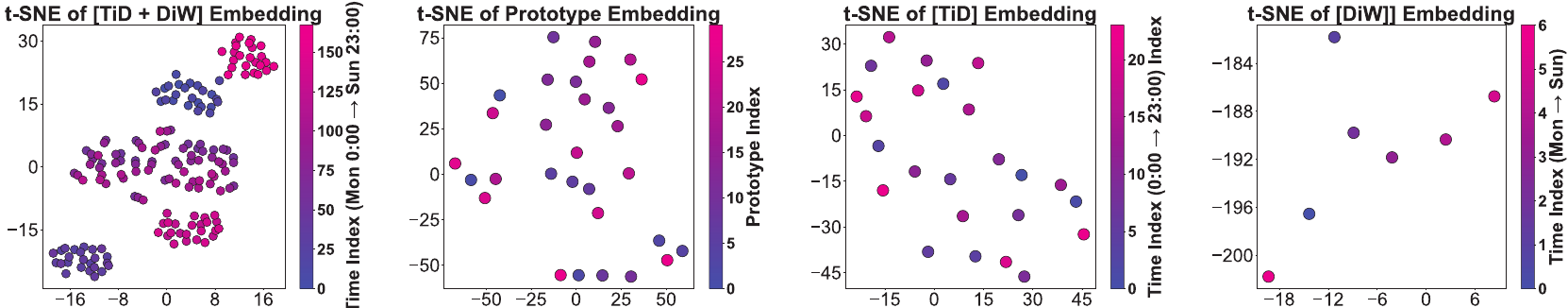}
									\caption{Visualization of APT's embeddings on ETTh1 dataset and iTransformer}
									\label{tsneetth1}
								\end{figure*}
								\begin{figure*}[t]
									\centering
									\includegraphics[width=0.92\textwidth]{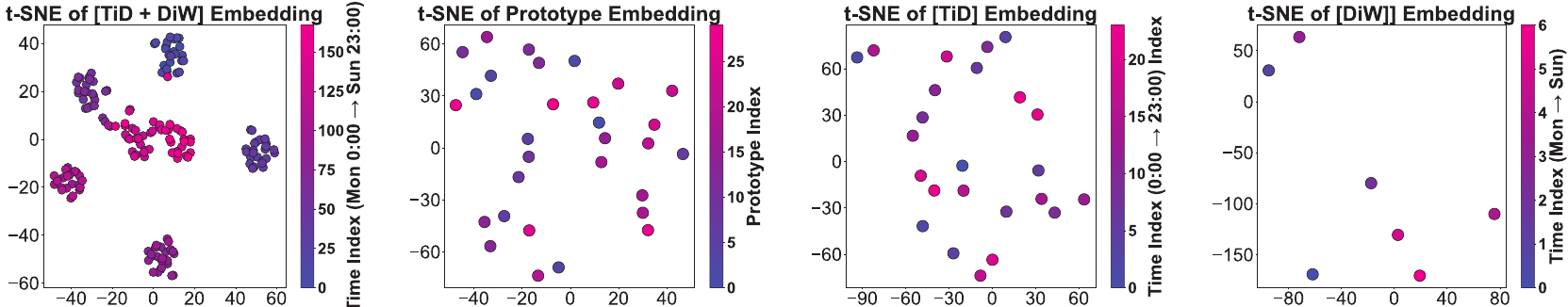}
									\caption{Visualization of APT's embeddings on ETTh2 dataset and iTransformer}
									\label{tsneetth2}
								\end{figure*}
								\begin{figure*}[t]
									\centering
									\includegraphics[width=0.46\textwidth]{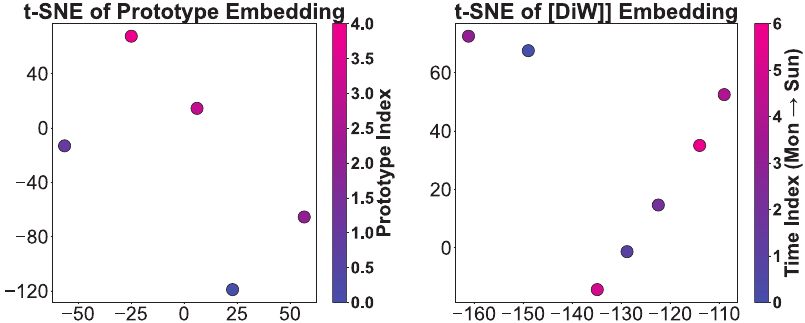}
									\caption{Visualization of APT's embeddings on Exchange dataset and iTransformer}
									\label{tsneexchangerate}
								\end{figure*}
								\begin{figure*}[t]
									\centering
									\includegraphics[width=0.92\textwidth]{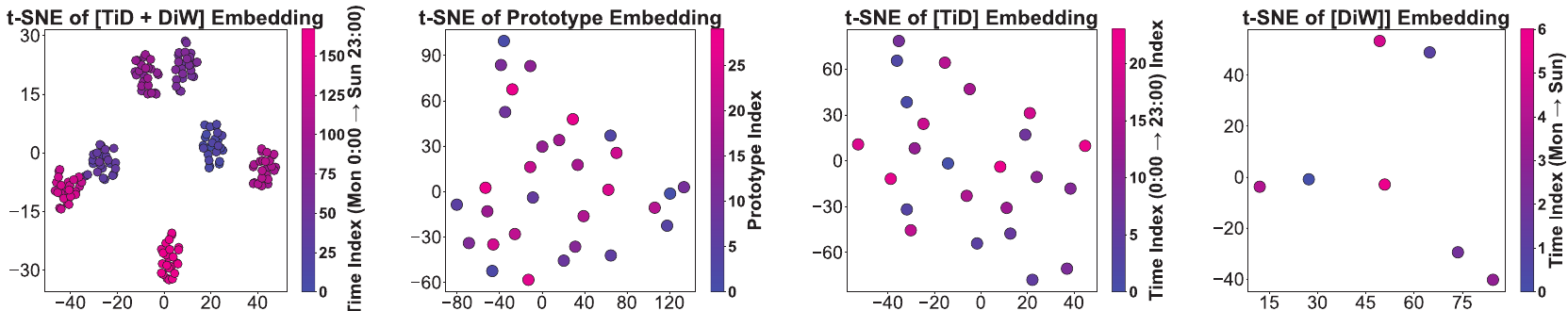}
									\caption{Visualization of APT's embeddings on Traffic dataset and iTransformer}
									\label{tsnetraffic}
								\end{figure*}
								\begin{figure*}[t]
									\centering
									\includegraphics[width=0.92\textwidth]{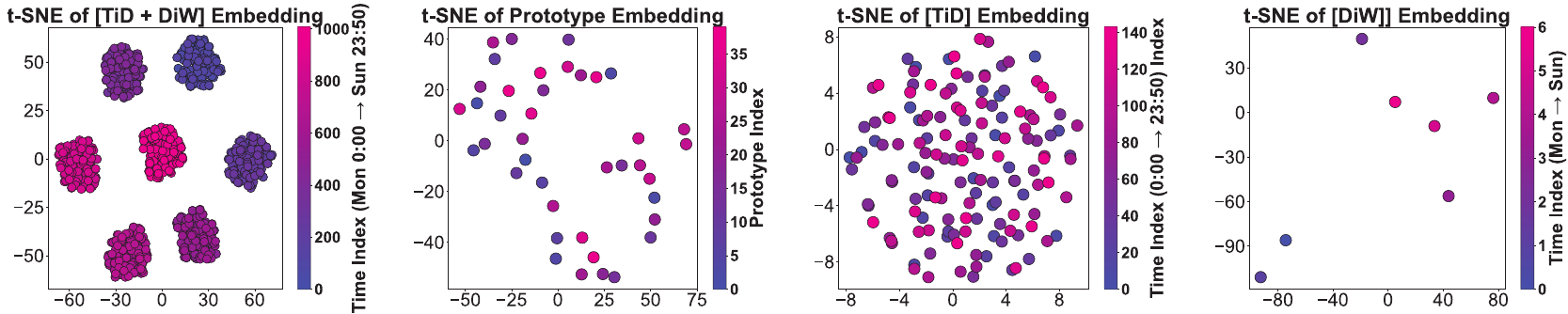}
									\caption{Visualization of APT's embeddings on Weather dataset and iTransformer}
									\label{tsneweather}
								\end{figure*}
								
								\begin{figure*}[t]
									\centering
									\includegraphics[width=0.95\textwidth]{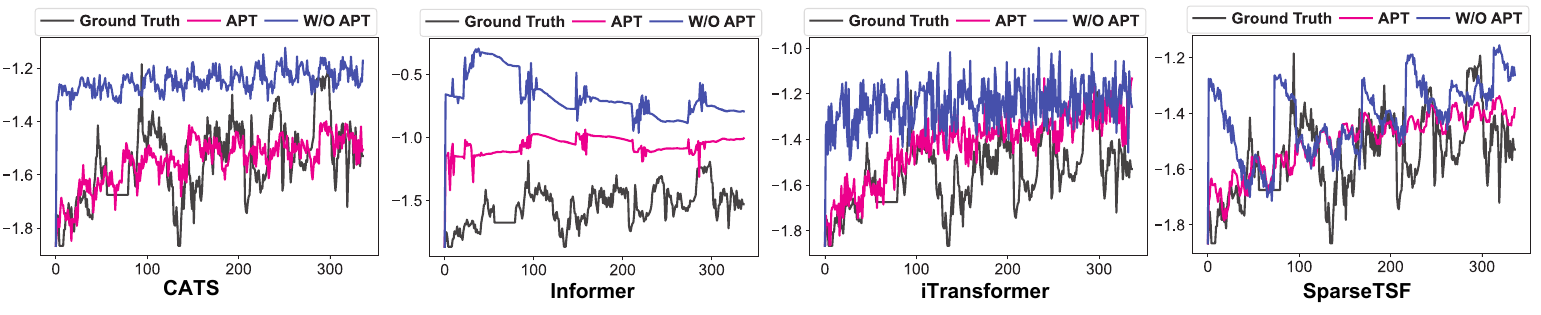}
									\caption{Visualization of forecasting results for different models on the ETTh1 dataset without normalization strategy}
									\label{caseetth1}
								\end{figure*}
								\begin{figure*}[t]
									\centering
									\includegraphics[width=0.95\textwidth]{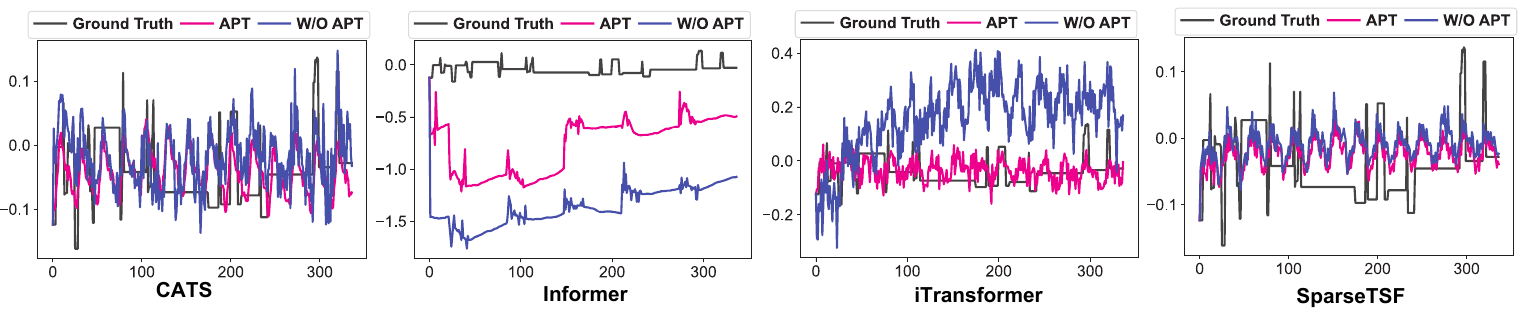}
									\caption{Visualization of forecasting results for another channel on the ETTh2 dataset without normalization strategy for different models}
									\label{caseetth22}
								\end{figure*}
								\begin{figure*}[t]
									\centering
									\includegraphics[width=0.95\textwidth]{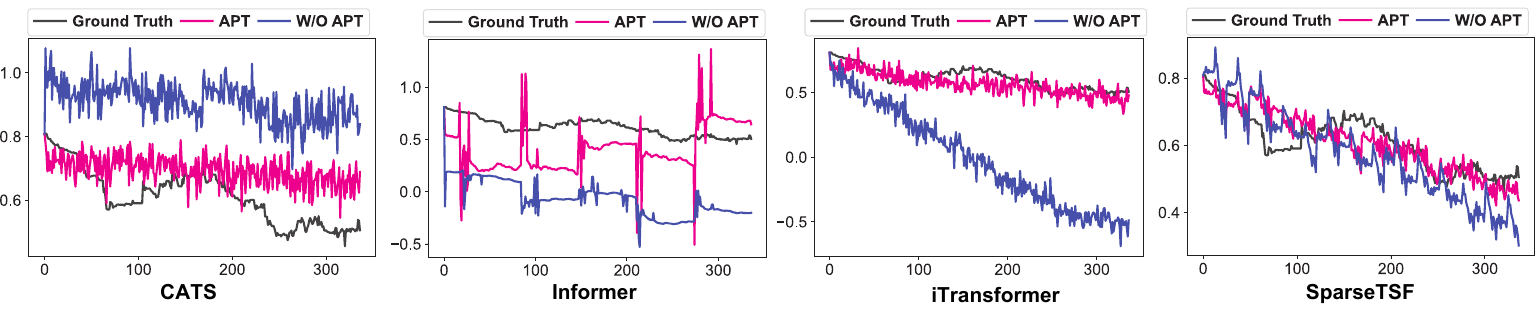}
									\caption{Visualization of forecasting results for different models on the ExchangeRate dataset without normalization strategy}
									\label{caseex}
								\end{figure*}
								\begin{figure*}[t]
									\centering
									\includegraphics[width=0.90\textwidth]{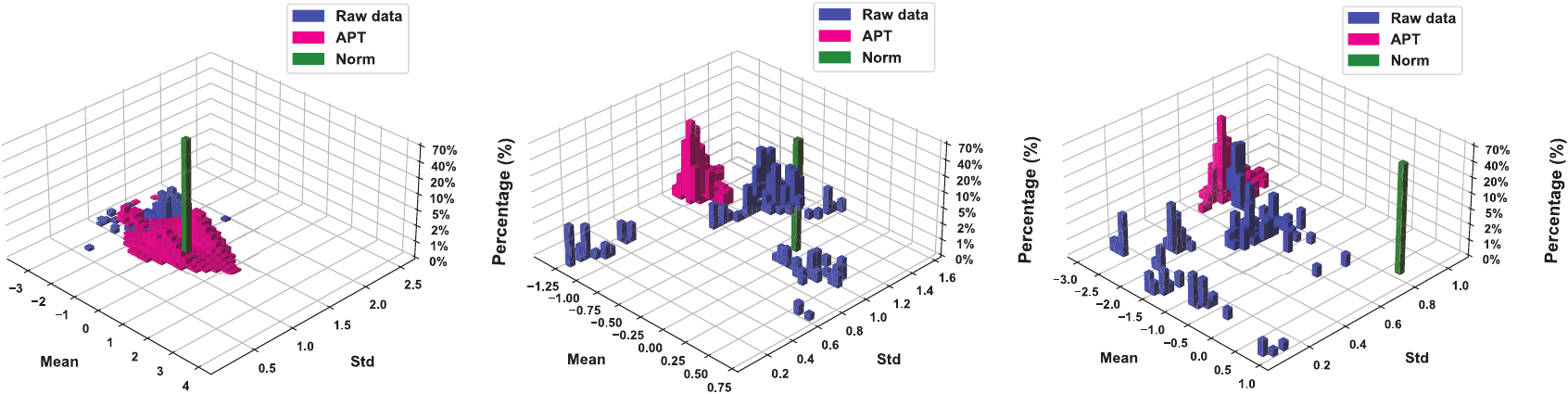}
									\caption{3D visualization of temporal distribution and ratio of forecasting pipeline at different stages on ELC, ETTh1 \& ETTh2.}
									\label{dis1}
								\end{figure*}
								\begin{figure*}[t]
									\centering
									\includegraphics[width=0.90\textwidth]{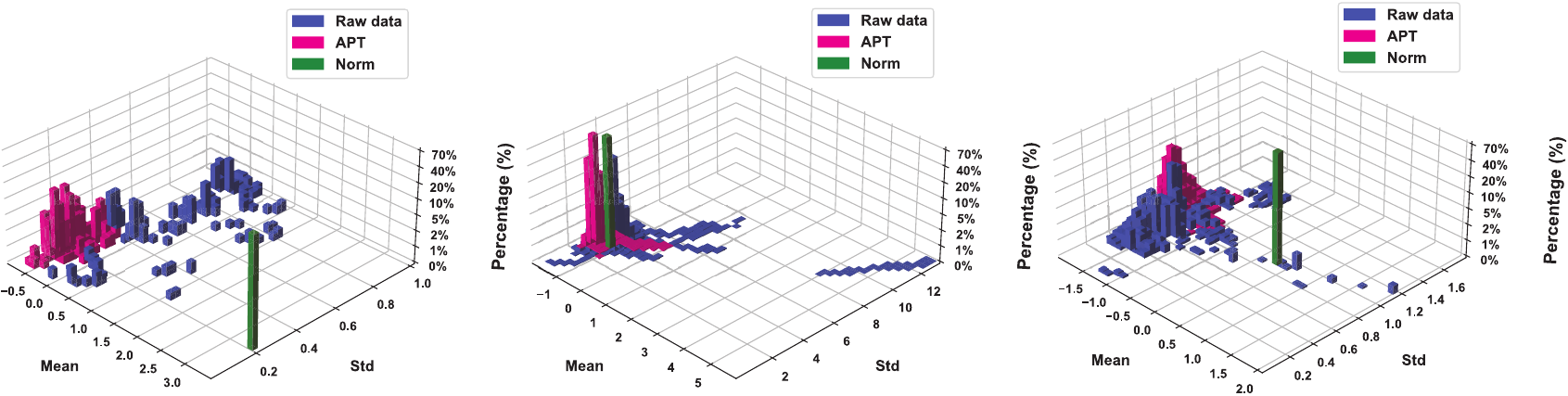}
									\caption{3D visualization of temporal distribution and ratio of forecasting pipeline at different stages on Exchange, Traffic \& Weather.}
									\label{dis2}
								\end{figure*}
								\begin{figure*}[t]
									\centering
									\includegraphics[width=0.9\textwidth]{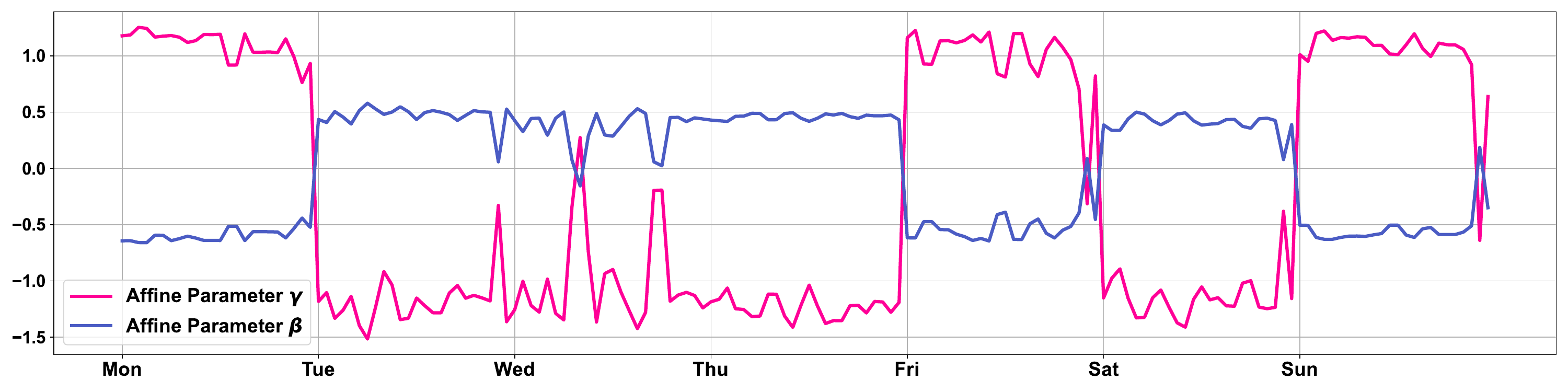}
									\caption{Visualization of affine parameters at different timestamps over a week on the ELC datasets}
									\label{affineElectricity}
								\end{figure*}
								\begin{figure*}[t]
									\centering
									\includegraphics[width=0.9\textwidth]{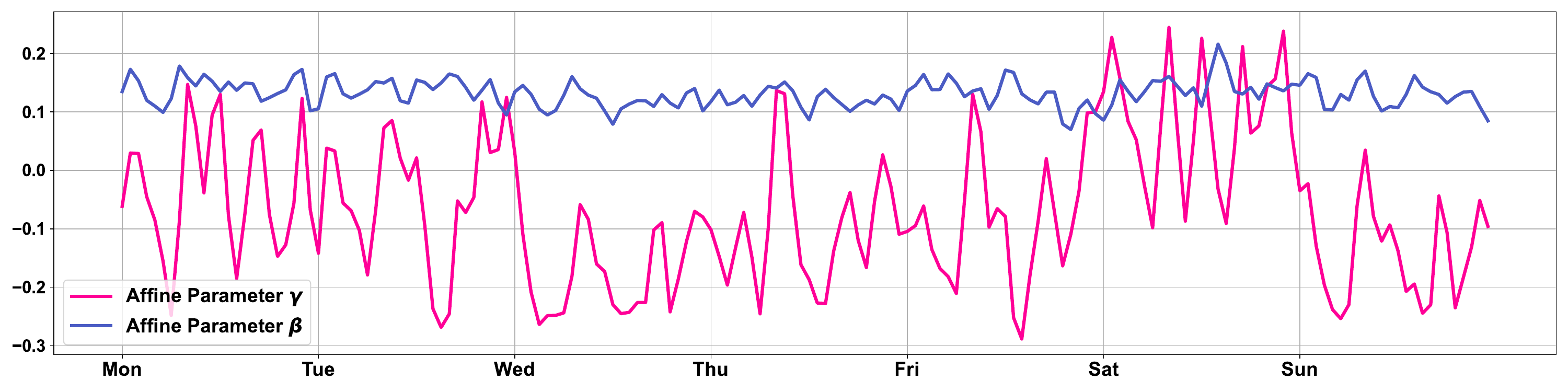}
									\caption{Visualization of affine parameters at different timestamps over a week on the ETTh1 datasets}
									\label{affineetth1}
								\end{figure*}
								\begin{figure*}[t]
									\centering
									\includegraphics[width=0.9\textwidth]{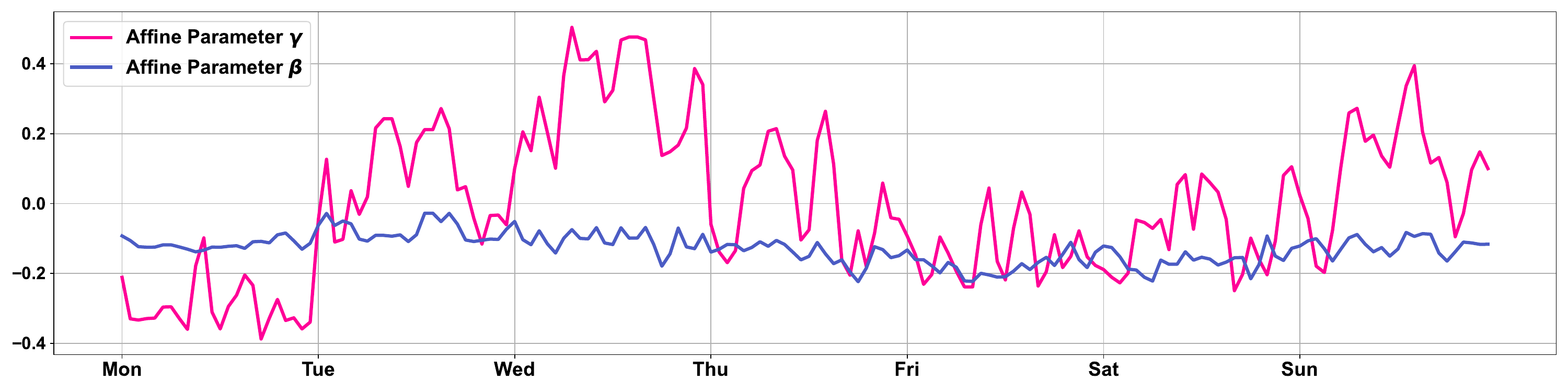}
									\caption{Visualization of affine parameters at different timestamps over a week on the ETTh2 datasets}
									\label{affineetth2}
								\end{figure*}
								\begin{figure*}[t]
									\centering
									\includegraphics[width=0.9\textwidth]{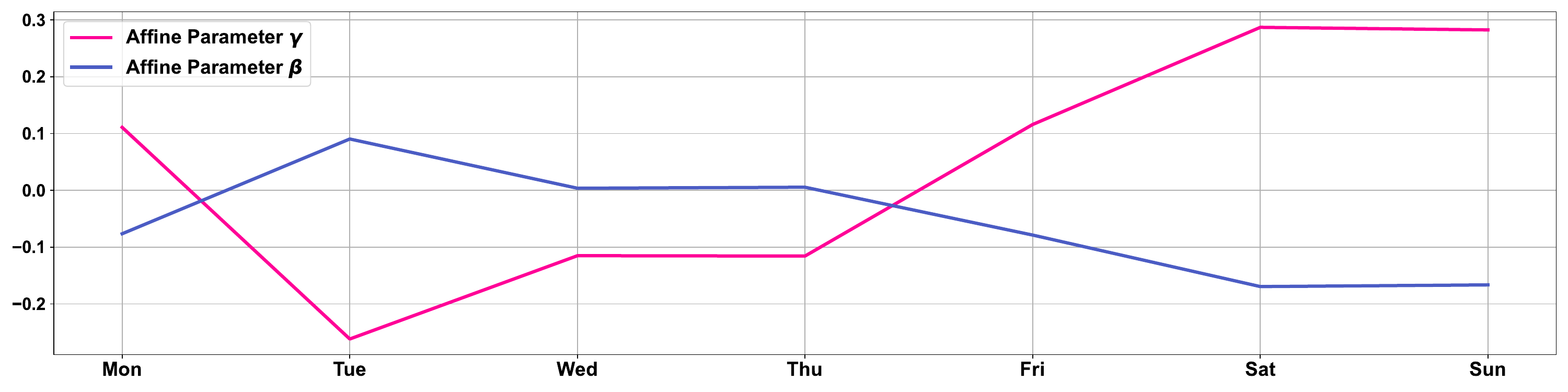}
									\caption{Visualization of affine parameters at different timestamps over a week on the Exchange datasets}
									\label{affineExchangeRate}
								\end{figure*}
								\begin{figure*}[t]
									\centering
									\includegraphics[width=0.9\textwidth]{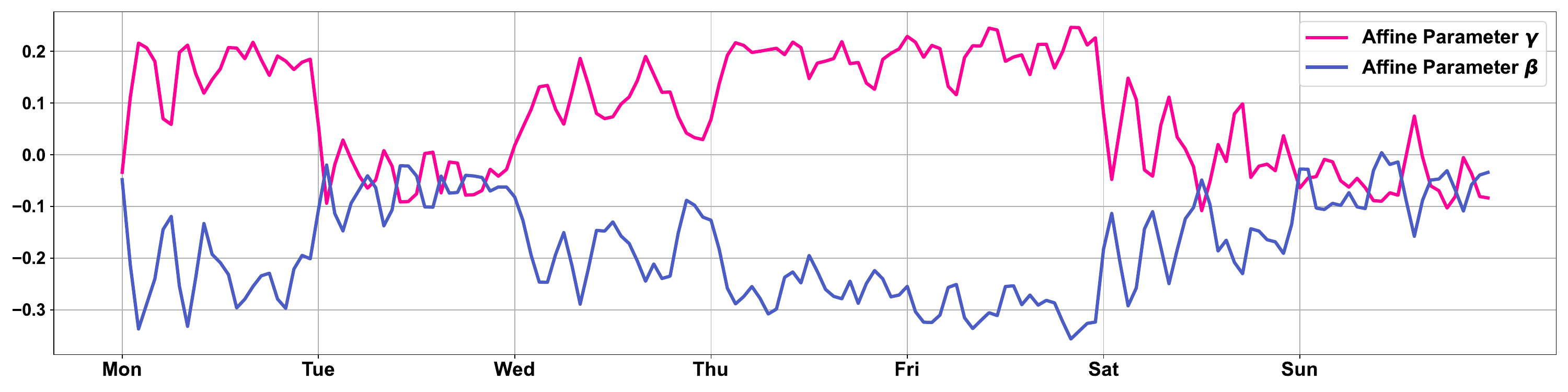}
									\caption{Visualization of affine parameters at different timestamps over a week on the Traffic datasets}
									\label{affineTraffic}
								\end{figure*}
								\begin{figure*}[t]
									\centering
									\includegraphics[width=0.9\textwidth]{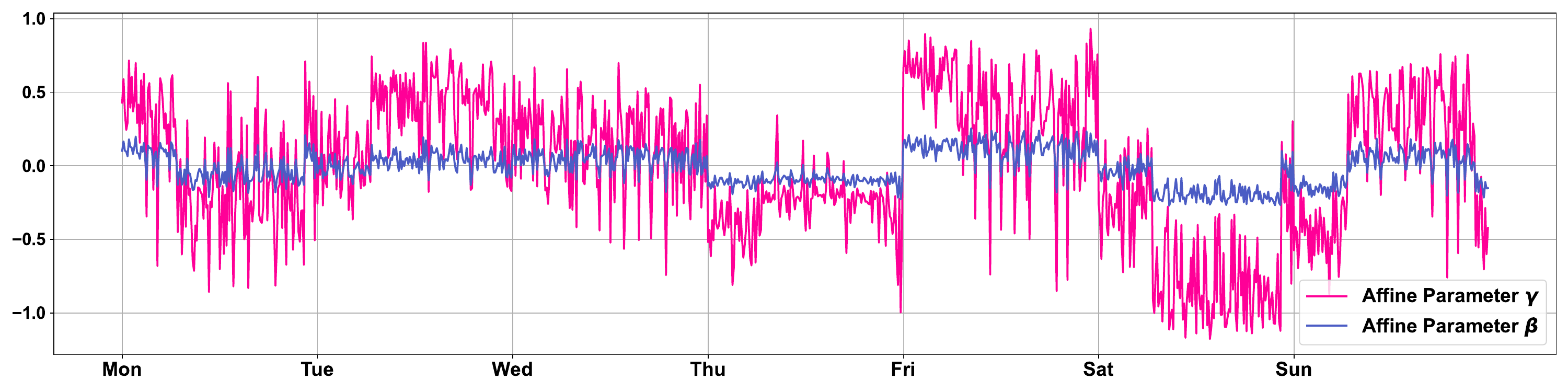}
									\caption{Visualization of affine parameters at different timestamps over a week on the Weather datasets}
									\label{affineWeather}
								\end{figure*}

								\subsection{Visualization}\label{visual}
								\subsubsection{t-SNE of Embedding:} Figure~\ref{tsneecl}-\ref{tsneweather} shows the timestamp and prototype embeddings learned by APT with iTransformer on all datasets. 
								In general, timestamp representations tend to cluster around “Day in Week”, except on ETT1/2, where human behavioral patterns may introduce additional coupling.
								Prototypes, “Day in Week”, and “Time in Day” are inherently discrete. However, due to the soft nature of the orthogonality loss $Loss_{orth}$ and the influence of external factors across datasets, their representations still exhibit structured patterns in the t-SNE projected space.
								\subsubsection{Forecasting cases:} Figure~\ref{caseetth21} and Figure~\ref{caseetth1}-\ref{caseex} presents forecasting examples on the ETTh1, ETTh2 and Exchange datasets without normalization.  In these dataset cases, backbone models often exhibit severe distributional shift. For example, Informer forecastings generally have a clear mean offset from the ground truth, while iTransformer outputs reveal systematic trend bias.
								
								APT effectively mitigates these issues using only two-dimensional dynamic affine parameters per instance. As shown in Figure~\ref{caseetth21}, APT amplifies periodic scales; In Figure~\ref{caseetth22}, it reduces volatility of pulse-like time series; In Figures~\ref{caseetth1} and \ref{caseex}, it corrects deviations in both mean and trend.
								
								These cases collectively demonstrate APT’s mechanism for handling distribution shift, which adjusts scale and mean to better align predicted outputs with real-world distributions, without altering the underlying temporal patterns. This enables seamless compatibility between APT and a wide range of backbones.
								\subsubsection{Temporal distribution:}As illustrated in Figure~\ref{dis1} and ~\ref{dis2}, we visualize the mean, variance, and their ratio of the raw time series, RevIN-normalized sequences, and APT-transformed outputs in a 3D distribution space. All datasets exhibit varying degrees of distributional shift. The raw data differ significantly in distribution across datasets, with ETT and Exchange showing the most severe shift, followed by Traffic and Weather, while ECL remains relatively stable, which is consistent with our earlier dataset analysis.
								
								RevIN maps all sequences into a standardized space with zero mean and unit variance. While this prevents the backbone from being overwhelmed by distributional complexity, it also eliminates the possibility of leveraging meaningful distributional cues.
								In contrast, APT strikes a balance between smoothing and diversity through learnable affine transformations. The transformed sequences form clusters in distributional space—not overly uniform, but structured and distinct—allowing the backbone to perceive informative distributional patterns.
								
								These 3D visualizations consistently validate that APT operates on distributional alignment: by introducing global distributional awareness and controlled feature clustering, it can effectively mitigate distribution shift across a wide range of scenarios.
								\subsubsection{Affine Transformation:} Since our affine parameters are dynamically generated from timestamp information, we can visualize them corresponding to the entire “Time in Day” and “Day in Week” timestamps on the test set. Figure~\ref{affineElectricity}-\ref{affineWeather} presents the learned weekly affine parameters from APT on all datasets using iTransformer as the backbone. Most datasets are sampled hourly, except for Exchange~(daily) and Weather~(every 10 minutes). Since these parameters are derived from deep networks, their distributions are inherently difficult to interpret in a fully human-interpretable manner.
								
								Nonetheless, we observe that most datasets exhibit distinct affine patterns across different “Day in Week”, which is consistent with the previous t-SNE visualization of embeddings. Importantly, the clustered affine parameters remain bounded, which is critical for convergence in regression tasks. This constraint is guided by our Affine Regularization Loss $Loss_{R}$, which softly encourages the parameters to maintain zero mean and unit variance. However, due to the soft nature of self-supervision, the loss primarily enforces boundedness rather than strict normalization.
								
								Interestingly, datasets like ECL, Exchange, and Traffic exhibit axis-symmetric patterns, which we speculate result from the scale parameter $\gamma$ overfitting to redundant distributional signals, with the bias $\beta$ compensating to restore balance.
								
								\section{Discussion}~\label{discussion}
								\subsection{Pattern, Distribution \& Distribution Shift}
								We aim to establish a shared understanding of the core components of time series data—pattern and distribution.
								Patterns refer to the structural dynamics of the series, such as combinations of trend, seasonality, and residuals. Forecasting backbones are primarily designed to capture these patterns, as they reflect both short- and long-term dependencies critical to accurate prediction.
								
								Distributions, on the other hand, describe the state space of the series. A classic perspective views distributions through statistical properties such as mean and variance, often under the assumption of Gaussianity. Since distributions are shaped by external conditions, they tend to drift over time, posing a fundamental challenge. When distributional shift are entangled with pattern dynamics, the value ranges and variability observed by the backbone vary significantly. This forces the model to allocate capacity toward fitting distributional noise, often at the cost of learning meaningful patterns.
								
								One of the dominant strategies to address distribution shift is normalization. By mapping sequences into a common statistical space~(e.g., zero mean and unit variance), normalization enables the backbone to focus solely on pattern learning. A subsequent denormalization step restores the output to its original scale. This design simplifies the learning problem and enhances performance.
								Recent works such as Dish-TS, SAN, and FAN extend the simple RevIN framework by focusing on the learnable components of distributions like modeling asymmetric statistics between history and future, or going beyond mean-variance assumptions.
								
								\subsection{Local statistical property and global temporal semantics}
								While statistical normalization with local mean and variance is widely adopted, it has inherent limitations. First, time series often contain noise and missing values, making such statistics unreliable. Second, these statistics are computed over limited contexts: RevIN relies on full-length history, while SAN uses patch-level features, yet neither guarantees alignment with the actual distribution of future samples. Furthermore, relying solely on Gaussian assumptions~(mean \& variance) may fail to capture more comprehensive distribution features.
								
								To overcome these limitations, we argue for leveraging global, non-statistical information. APT adopts timestamps as external priors to mitigate local errors. Timestamps are easy to obtain and strongly correlated with real-world temporal semantics such as traffic spikes during rush hours, high TV ratings at night, or increased foot traffic in shopping malls on weekends. As a result, they implicitly encode global information that statistical normalization alone cannot capture.
								
								\subsection{Reasons for affine transformation}
								APT utilizes a network to apply dynamic affine parameters to time series with different timestamp combinations. Affine transformations have long been integrated with normalization. For instance, in Transformers to align intermediate representations between layers or blocks. More significantly, AdaIN in style transfer showed that affine parameters can serve as carriers of cross-modal information, which demonstrates that such parameters can effectively encode external priors to assist a primary task.
								
								Inspired by this, APT uses affine modulation to inject timestamp-driven global distributional awareness into the forecasting pipeline. Unlike prior works such as GLAFF or Informer's temporal embedding, which encode timestamp features as full-sequence tensors with the same shape as the time series and use them either jointly or as replacements for the sequence itself, APT follows a minimal design. For example, GLAFF introduces more parameters than some backbones like CATS, and suffers from the uniqueness of timestamp combinations, which it difficult to extract generalizable patterns, similar to few-shot learning challenges.
								
								APT addresses this by producing only two parameters per instance and applying them to modulate the sequence through affine transformation. This lightweight design avoids redundancy, minimizes overhead, and, with the help of prototype learning, ensures robust and scalable global timestamp representation.
								
								\subsection{Theoretical support for APT}
								Generally speaking, the work related to conditional affine parameters in computer vision and natural language processing does not require theoretical explanations because the ideas are very simple. Introducing external variables such as timestamps is equivalent to introducing conditional information into the modeling of time series. According to the principles of Bayesian inference and information gain, conditioning on informative variables reduces uncertainty, thereby improving generalization and forecasting accuracy.
								
								While the MSE loss used in point forecasting does not strictly follow a Bayesian formulation, we provide the following conceptual argument to clarify APT's benefit:
								
								In time series forecasting, the goal is to estimate the conditional probability of the future series $Y$ given the historical sequence $X$, denoted as $P(Y\mid X)$. By treating timestamp $ts$ as an informative variable, we shift to estimating:
								\[
								P(Y \mid X, ts) = \frac{P(ts \mid Y, X), P(Y \mid X)}{P(ts \mid X)}
								\]
								This conditioning provides additional benefit to forecasting. According to the conditional entropy relation:
								\[
								\mathcal{H}(Y \mid X, ts) \leq \mathcal{H}(Y \mid X),
								\]
								the inclusion of $ts$ always reduces or maintains the uncertainty in $Y$, leading to improved forecasting performance. We believe that this conditional information can actually be replaced by any task-related information, such as text or images, which is consistent with the broader idea of multimodal and multivariate time series forecasting.
								
								\section{Limitation and Future Work}
								\subsubsection{Limitation:} APT relies on manually chosen timestamps such as "Time in Day" or "Day in Week" with distributional similarity as shown in Figure~\ref{js} to obtain information gain. However, in the appendix experiments, irrelevant labels "Day in Month" not only fail to provide performance gains but may even degrade forecasting accuracy. 
								
								APT requires pretraining with additional losses. Moreover, since APT relies on self-supervised clustering and requires parameter alignment with the forecasting task, improper hyperparameter configurations may hinder its ability to enhance the performance of the backbone model, posing challenges for practical deployment.
								
								\subsubsection{Future Work:} APT introduces a new paradigm in time series forecasting by addressing the limitations of local statistical normalization and mitigating distribution shift. Future directions include integrating online distribution shift detection, incorporating richer global temporal semantics, and exploring more expressive parametric transformations.
								
								Moreover, we hope APT inspires multimodal research in time series.  Timestamps are sparse yet semantically rich signals, often requiring large models like GLAFF for effective representation before our work. However, each time series sample typically receives only limited external context.  Unlike vision or language data, 1-D time series usually cannot accommodate unique but semantically rich external information, often resulting in overfitting of multimodal tasks of time series. APT leverages dynamic affine transformations to compress such redundant, noisy inputs into compact parametric forms, which serves as an information attenuator that aligns external modalities with the limited expressiveness of time series. We consider that this technology is expected to help better integrate  auxiliary signals and enhances performance in cross-modal forecasting tasks.

\end{document}